\documentclass[journal]{IEEEtran}
\usepackage{cite}
\usepackage{booktabs}
\usepackage{amsmath}
\usepackage{mathrsfs}
\usepackage{ntheorem}
\usepackage{extarrows}
\usepackage[ruled]{algorithm2e}
\usepackage{makecell}
\usepackage{multirow}
\usepackage{color}
\usepackage{threeparttable}
\usepackage{amssymb}
\makeatletter
\theoremseparator{:}
\newtheorem{myDef}{\it Definition}
\newtheorem{myProperty}{\it Property}
\newtheorem{myRemark}{\it Remark}
\newtheorem*{myProof}{\it Proof}
\renewcommand{\maketag@@@}[1]{\hbox{\m@th\normalsize\normalfont#1}}
\makeatother

\ifCLASSINFOpdf
  \usepackage[pdftex]{graphicx}
  \graphicspath{{../pdf/}{../jpeg/}}
  \DeclareGraphicsExtensions{.pdf,.jpeg,.png}
\else
  \usepackage[dvips]{graphicx}
  \graphicspath{{../eps/}}
  \DeclareGraphicsExtensions{.eps}
\fi

\begin{document}
%
\title{Perching on Moving Inclined Surfaces using Uncertainty Tolerant Planner and Thrust Regulation}

%
%
%

\author{Sensen~Liu, 
        Wenkang~Hu,
        Zhaoying~Wang,
        Wei~Dong,~\IEEEmembership{Member,~IEEE,}
        and Xinjun~Sheng,~\IEEEmembership{Member,~IEEE}
\thanks{This work is partially supported by the National Natural
Science Foundation of China (Grant No. 51975348) and
Shanghai Rising-Star Program(22QA1404400). \textit{(Corresponding author: Wei Dong and Xinjun Sheng.)}}
\thanks{The authors are with the State Key Laboratory
of Mechanical System and Vibration, Robotics Institute, School of Mechanical and Engineering, Shanghai Jiao Tong University, 800
Dongchuan Road, Shanghai, China. \tt\footnotesize (e-mail: sensenliu@sjtu.edu.cn; chn.h.w.k@sjtu.edu.cn;wangzhaoying@sjtu.edu.cn; dr.dongwei@sjtu.edu.cn;xjsheng@sjtu.edu.cn).}}
%



\maketitle

\begin{abstract}
Quadrotors with the ability to perch on moving inclined surfaces can save energy and extend their travel distance by leveraging ground vehicles. Achieving dynamic perching places high demands on the performance of trajectory planning and terminal state accuracy in SE(3). However, in the perching process, uncertainties in target surface prediction, tracking control and external disturbances may cause trajectory planning failure or lead to unacceptable terminal errors. To address these challenges, we first propose a trajectory planner that considers adaptation to uncertainties in target prediction and tracking control. To facilitate this work, the reachable set of quadrotors' states is first analyzed. The states whose reachable sets possess the largest coverage probability for uncertainty targets, are defined as optimal waypoints. Subsequently, an approach to seek local optimal waypoints for static and moving uncertainty targets is proposed. A real-time trajectory planner based on optimized waypoints is developed accordingly. Secondly, thrust regulation is also implemented in the terminal attitude tracking stage to handle external disturbances. When a quadrotor's attitude is commanded to align with target surfaces, the thrust is optimized to minimize terminal errors. This makes the terminal position and velocity be controlled in closed-loop manner. Therefore, the resistance to disturbances and terminal accuracy is  improved. Extensive simulation experiments demonstrate that our methods can improve the accuracy of terminal states under uncertainties. The success rate is approximately increased by $50\%$ compared to the two-end planner without thrust regulation. Perching on the rear window of a car is also achieved using our proposed heterogeneous cooperation system outdoors. This validates the feasibility and practicality of our methods.
\end{abstract}


%
\IEEEpeerreviewmaketitle

\section{Introduction}
\IEEEPARstart{R}{ecent} advances in Micro Aerial Vehicles (MAVs)  perch signal a positive trend toward perching on moving objects\cite{ji2022real,zhang2020fast,vlantis2015quadrotor}. MAVs, specifically quadrotors, can significantly extend their mission time and landing points with this capability\cite{choudhury2021efficient,hsiao2022novel}. One category of these researches is investigating the techniques to aggressively perch on moving inclined surfaces. The surfaces on ground vehicles can be thus exploited\cite{lasla2019exploiting,baca2019autonomous}. Unfortunately, it is challenging to perch on moving inclined surfaces outdoors for quadrotors. It requires high accuracy of terminal state relative to moving targets in SE(3). The high accuracy relies on real-time trajectory planning and closed-loop control. However, the uncertainties in predicted targets and tracking control will lead to failure of real-time planning. The closed-loop control of position and velocity is also ignored when the attitude is commanded to align with the target inclined surface. All these factors can increase terminal pose errors and eventually lead to a failed perch. The uncertainties and disturbances may be more significant in the outdoor environment. That hinders the practical application of MAVs for perching outdoors. 
\subsection{Related Work}
Research efforts have been mainly dedicated to decision-planning and control fields to handle uncertainties\cite{dadkhah2012survey}. In terms of planning, Brault et al.\cite{brault2021robust} propose a trajectory planner by reducing state and input sensitivity w.r.t. uncertainties in model parameters. To deal with uncertainties in prediction, the Partially Observable Markov Decision Process (POMDP) has been extensively exploited\cite{vermaelen2020survey,tao2021path}. Authors in\cite{bey2021handling,ding2021epsilon} employ POMDPs to model the uncertainties in moving traffic participants. Safe and comfortable rides are achieved. This method suffers from the curse of dimensionality and high computational complexity. It is appropriate for coarse  behavior planning. Another way is to formulate the uncertainties as continuous random variables to construct programming problems. Zhu and Xu et al.\cite{zhu2019chance,xu2021dpmpc} employ a linear Gaussian model to characterize the uncertainties in dynamics and obstacle motion. A chance-constrained nonlinear model predictive control (CCNMPC) is constructed to plan collision avoidance trajectories among moving obstacles. It is then improved using non-Gaussian distributions to represent the uncertainties\cite{wang2020non,zhang2020trajectory}. Although it performs well in dynamic collision avoidance, CCNMPC is still computationally intensive or conservative with low accuracy. 

Uncertainties in the future states of the dynamic system are also addressed by reachability analysis (RA) to design robust and safe trajectory planners\cite{gueguen2009safety}. The trajectory-dependent uncertainties induced by tracking errors are described by a Forward Reachable Set (FRS) in \cite{kousik2019safe}. A safe trajectory planner based on the reachability analysis is developed to achieve aggressive flight in simulated cluttered environments. Manzinger et al. \cite{manzinger2020using} adopt RA to obtain driving corridors in complex road scenarios. Online trajectory planning is implemented in convoluted spaces. The authors in \cite{lakhal2022safe} analyse the vehicle's reachability when it moves towards sparse waypoints. Different uncertainties in modeling are considered and propagated to obtain reachable sets. An adaptive control law is developed to avoid the intersection of reachable sets and obstacles. Safe navigation is thus achieved using sparse waypoints. Although the concept of RA has been widely used in guaranteed-safe planning, the potential of adapting to uncertainties in moving targets still needs to be further tapped.

In trajectory tracking control, various feedback control strategies are also proposed to alleviate the influence of uncertainties in dynamic systems and external disturbances \cite{yang2021learning,tran2019adaptive,wang2022integrated,lu2020uncertainty}. Many of them are focused on tracking three-dimensional position and yaw angle, called flat outputs. The non-flat outputs, the roll and pitch angles, depend on the position trajectory tracking controller due to the underactuation of quadrotors. This cannot cater to the requirements for perching on an inclined surface where attitudes and positions must be independently controlled. 

To circumvent the underactuation limitations, a discrete-time non-linear model predictive controller (NMPC) is constructed to optimize the control input of a quadrotor. The landing on a moving inclined platform is realized\cite{vlantis2015quadrotor,hu2019time,sun2021comparative}. The underactuation model is considered in the MPC problem such that the discrepancy between position and attitude tracking is avoided. This method costs excessive computation time and the slope of the inclined surface is also relatively small in their works. To decouple the requirements of the position and attitude tracking, the authors in \cite{pi2021reinforcement,liu2022hitchhiker} propose a two-stage control strategy and achieve perching on static and moving vertical surfaces. The strategy contains position-velocity tracking and attitude tracking. If the distance of quadrotors to target surfaces is large, position-velocity tracking is employed to track three-dimensional positions and their derivatives. Otherwise, attitude tracking is used to control the quadrotor's attitude to be consistent with the surface's orientation. This concise scheme is also used to achieve aggressive flying through multiple inclined openings\cite{guo2020image}. In existing studies, during the attitude tracking stage, the attitude is determined by target orientation, and thrust is artificially specified. The positions and velocities are controlled in an open-loop manner. This causes their accuracies to be susceptible to disturbances. Therefore, the closed-loop control of position and velocity during the attitude tracking stage still needs further study.

\subsection{Contributions of This Article}
To address the issues above, we propose a trajectory planner adapting uncertainties in  target predictions and tracking control as well as an optimal controller to regulate thrust in the attitude tracking stage. As for the trajectory planner, the forward reachable sets of candidate waypoints are analyzed. Their coverage probability for the distribution of predicted targets, considering tracking errors, is calculated. Then, sequential optimized waypoints are selected to maximize their expected coverage probability. The transformation rule of optimized waypoints w.r.t. different targets are derived, enabling the waypoints to be used in real-time planning. As regards the thrust optimization control, we design a model predictive control strategy to regulate the thrust. The thrust can be exploited to minimize the terminal position and velocity errors. To make the problem tractable, we transform the original problem into a nonlinear programming problem by using switching times as decision variables. Then, the optimal thrust can be attained. The positions and velocities can be controlled in closed-loop. We also design a heterogeneous cooperation framework to verify our proposed method using a master hexacopter for detection-command and a slave quadrotor for execution. A series of comparison simulations are implemented to verify the effectiveness of the proposed method. The experiments where a quadrotor perches on the rear window of a moving car in outdoor environments are also conducted using the cooperation framework. To the best of our knowledge, this is the first work achieving perching on a moving inclined surface (up to $70^\circ,2.15m/s$) outdoors.

The contributions of this work are threefold:

1) A trajectory planner based on waypoint optimization using reachability analysis is proposed to adapt to uncertainties in target predictions and tracking control.

2) An optimal controller to regulate thrust in the attitude tracking stage is developed to alleviate influences of disturbances.

3) A series of comparison simulations and outdoor validation experiments are conducted using a devised heterogeneous cooperation framework to verify the proposed methods.  

This paper is structured as follows: In Section \uppercase\expandafter{\romannumeral2}, we describe the problem to be solved. In Section \uppercase\expandafter{\romannumeral3}, we introduce the planning method to adapt to uncertainties. The thrust optimization controller in the attitude tracking stage is presented in Section \uppercase\expandafter{\romannumeral4}. The experiments and corresponding results are demonstrated in Sections \uppercase\expandafter{\romannumeral5}. The conclusion is finished in Sections \uppercase\expandafter{\romannumeral6}.

\section{Problem Statement}
\begin{figure}[!tb]
  \centering
  \includegraphics[width=3.0in]{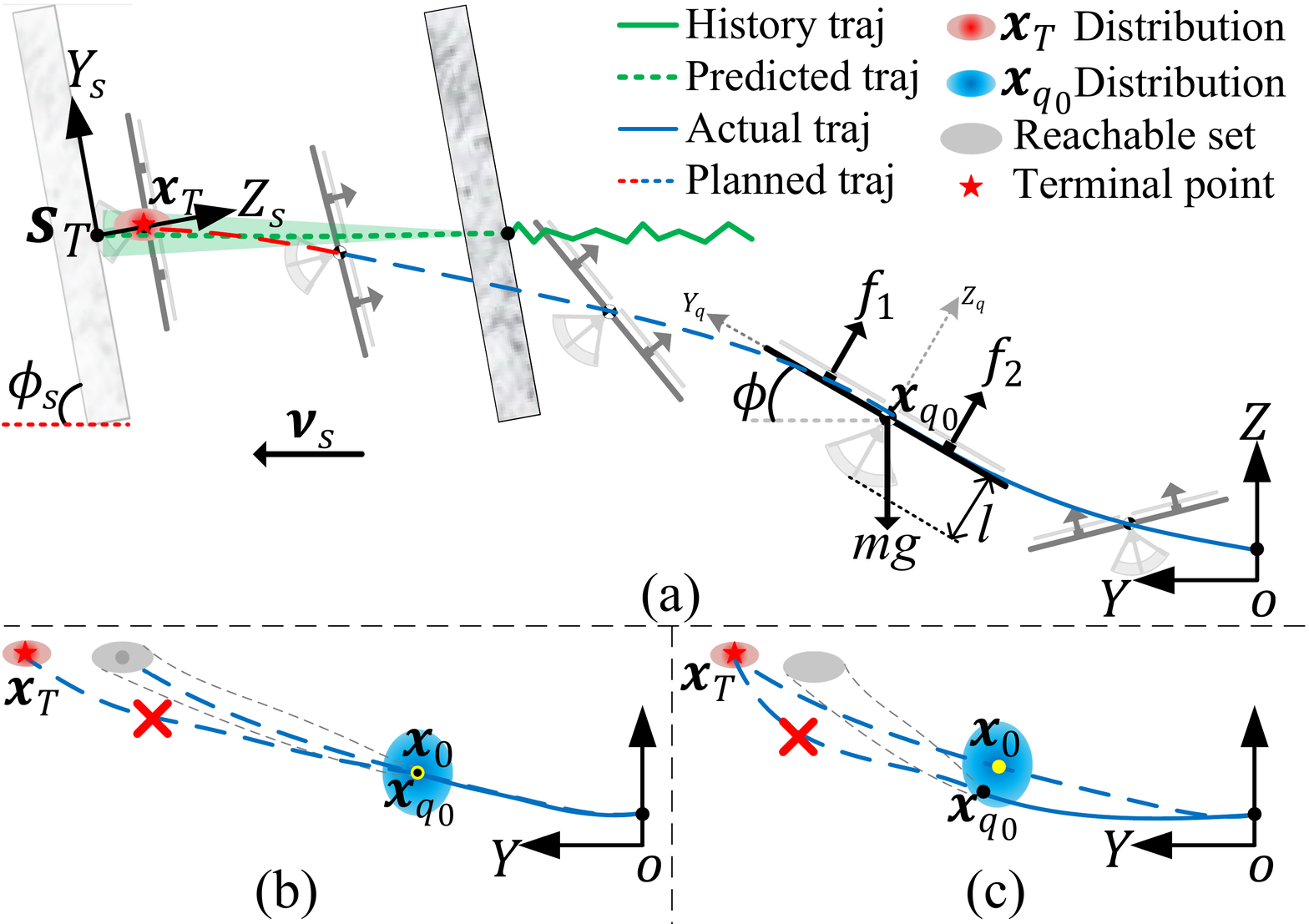}
  \caption{The process of a quadrotor perching on a moving inclined surface and two cases of planning failure. The green solid line is the history trajectory of the surface. The green dash line is the corresponding predicted future trajectory. The blue solid line is the real trajectory of the quadrotor. The blue and red dashed line is the planned trajectory. (a) The process of perching. (b) Planning failure due to change of targets. (c) Planning failure due to tracking errors.}
  \label{initialproblemdescribe}
\end{figure}
In this article, vectors and matrices are denoted with bold letters. Unless stated otherwise, all vectors in this article are column vectors. The process of a quadrotor perching on a moving inclined surface is shown in Fig.\ref{initialproblemdescribe}(a). The quadrotor starts from  hover state and flies towards the moving surface by tracking a real-time planned trajectory. The trajectory is defined as:
$$\boldsymbol{x}(t)=[\boldsymbol{p}(t)^\top,\boldsymbol{v}(t)^\top,\boldsymbol{a}(t)^\top]^\top \in \mathbb{R}^9.$$ 
$\boldsymbol{p}(t),\boldsymbol{v}(t),\boldsymbol{a}(t) \in \mathbb{R}^3 $ are position, velocity and acceleration profile, respectively.  $\boldsymbol{x}_0=[\boldsymbol{p}_0^\top,\boldsymbol{v}_0^\top,\boldsymbol{a}_0^\top]^\top$ and $\boldsymbol{x}_T=[\boldsymbol{p}_T^\top,\boldsymbol{v}_T^\top,\boldsymbol{a}_T^\top]^\top$ are start and terminal points of the trajectory, respectively. The state of a quadrotor is denoted as $\boldsymbol{x}_q=[\boldsymbol{p}_q^\top,\boldsymbol{v}_q^\top,\boldsymbol{a}_q^\top,\boldsymbol{q}_q^\top]^\top \in \mathbb{R}^{13}$. $\boldsymbol{q}_q \in \mathbb{R}^4$ is the attitude quaternion of a quadrotor. The state of the surface is defined by $\boldsymbol{s}=[\boldsymbol{p}_s^\top,\boldsymbol{v}_s^\top, \boldsymbol{Z}_s^\top]^\top \in \mathbb{R}^9$. The entries are position, velocity and normal direction of the surface, respectively. The predicted future trajectory is $\boldsymbol{s}(t)\in \mathbb{R}^9$ and the predicted rendezvous of the surface is $\boldsymbol{s}_T=[\boldsymbol{p}_{s_T}^\top,\boldsymbol{v}_{s_T}^\top, \boldsymbol{Z}_{s_T}^\top]^\top$. The $\boldsymbol{x}_T$ is determined as follows:
\begin{footnotesize}
  \begin{equation}
  \boldsymbol{p}_T\!=\! \boldsymbol{p}_{s_T} \!-\!l\boldsymbol{Z}_{s_T},\;
  \boldsymbol{v}_T\!=\! \boldsymbol{v}_{s_T} \! -\! v_n\boldsymbol{Z}_{s_T}+v_\tau\boldsymbol{Y}_{s},\;
  \boldsymbol{a}_T=g\boldsymbol{Z}_{s_T}\!-\!g\boldsymbol{e}_3.\label{eqn_terminaldetermined}
  \end{equation}
\end{footnotesize}
Where $l$ is the length of the quadrotor's centroid to the bottom. $\boldsymbol{Y}_s$ is the axis of the surface-fixing coordinate system (see Fig.\ref{initialproblemdescribe}(a)). $v_n$ and $v_\tau$ are the desired normal and tangential velocities
relative to the surface to achieve perch. $g$ is gravitational acceleration. The $\boldsymbol{x}_T$ is obtained from the predicted trajectory of surfaces according to (\ref{eqn_terminaldetermined}). The uncertainties of predicted target will thus be transferred to $\boldsymbol{x}_T$. The two scenarios of replanning failure are shown in Fig.\ref{initialproblemdescribe} (b) and (c). Fig.\ref{initialproblemdescribe} (b) shows failure caused by uncertainties of predicted targets. Uncertainties in prediction lead large changes in the new $\boldsymbol{x}_T$. There might be no feasible trajectory from $\boldsymbol{x}_0$ to the new $\boldsymbol{x}_T$, if the new  $\boldsymbol{x}_T$ is outside the reachable set of the quadrotor in specified time $T$. In another scenario, if current states deviate from the planned trajectory owing to tracking errors such that reachable set of $\boldsymbol{x}_0$ is unable to cover $\boldsymbol{x}_T$. The planner will also fail to find a feasible trajectory to $\boldsymbol{x}_T$ (see Fig.\ref{initialproblemdescribe} (c)). This inspires us to seek waypoints that possesses larger reachable set to cover possible terminal states. Additionally, the waypoints should be robust to the tracking errors. In this article, we assume the predicted $\boldsymbol{x}_T$ belongs to a multivariate Gaussian distribution $\mathcal{N}(\boldsymbol{x}_T,\boldsymbol{\Sigma}_T)$. The actual states of the quadrotor belongs to a multivariate Gaussian process $\mathcal{N}(\boldsymbol{x}_0,\boldsymbol{\Sigma}_0)$ when the quadrotor flies towards the states $\boldsymbol{x}_0$ that is on the planned trajectory. This  characterizes the tracking errors if a quadrotor flies towards $\boldsymbol{x}_0$\cite{van2011lqg}. The problem is to find sequential optimal waypoints to make the probabilities to cover the target as large as possible, if a quadrotor tracking a trajectory passing these optimized waypoints.

\section{Uncertainty Tolerant Planner}
There are three subsections to describe the planning method. The first subsection demonstrates the concept of optimal dexterous waypoint and its property. The second subsection presents a pipeline to attain local optimal waypoints. The last subsection shows the planning and replanning method using the obtained waypoints.

\subsection{Definition and Property of Optimal Dexterous Waypoints}
We will show three definitions to introduce the concept of the optimal dexterous waypoint (ODW).
\begin{myDef}
Let $\boldsymbol{F}$ and $\boldsymbol{G}$ be the given kinodynamics constraints (velocity, acceleration, actuator and body rate limitation) and geometrical constraints (position limitation). The $T$ reachable set of a state $\boldsymbol{x}_0$ is the collection of states $\boldsymbol{x}(\boldsymbol{p},\boldsymbol{v},\boldsymbol{a})$ that can be reached after time $T$ from $\boldsymbol{x}_0$, respecting  $\boldsymbol{F}$ and $\boldsymbol{G}$. It is noted as $ {^{\boldsymbol{F}}_{\boldsymbol{G}}}\boldsymbol{\mathcal{P}}(\boldsymbol{p}_0,\boldsymbol{v}_0,\boldsymbol{a}_0,T)={^{\boldsymbol{F}}_{\boldsymbol{G}}}\boldsymbol{\mathcal{P}}(\boldsymbol{x}_0,T)$. The corresponding control input set is ${^{\boldsymbol{F}}_{\boldsymbol{G}}}\boldsymbol{\mathcal{U}}(\boldsymbol{x}_0,T,t),t \in [0,T]$.
\label{reachablesetDef}
\end{myDef} 

This definition represents the reachable states from $\boldsymbol{x}_0$ after time $T$. There is a feasible trajectory with duration $T$ from $\boldsymbol{x}_0$ to $\boldsymbol{x}_T$ equivalents to $\boldsymbol{x}_T \in {^{\boldsymbol{F}}_{\boldsymbol{G}}}\boldsymbol{\mathcal{P}}(\boldsymbol{x}_0,T)$. The addition and subtraction of a reachable set with a vector $\boldsymbol{b} \in \mathbb{R}^9$ are defined as: 
\begin{equation}
  \begin{split}
{^{\boldsymbol{F}}_{\boldsymbol{G}}}\boldsymbol{\mathcal{P}}(\boldsymbol{x}_0,T)\oplus \boldsymbol{b}= \{\boldsymbol{x}+\boldsymbol{b}|\boldsymbol{x} \in {^{\boldsymbol{F}}_{\boldsymbol{G}}}\boldsymbol{\mathcal{P}}(\boldsymbol{x}_0,T),\boldsymbol{b}\in\mathbb{R}^9 \},\\
{^{\boldsymbol{F}}_{\boldsymbol{G}}}\boldsymbol{\mathcal{P}}(\boldsymbol{x}_0,T)\ominus \boldsymbol{b}= \{\boldsymbol{x}-\boldsymbol{b}|\boldsymbol{x} \in {^{\boldsymbol{F}}_{\boldsymbol{G}}}\boldsymbol{\mathcal{P}}(\boldsymbol{x}_0,T),\boldsymbol{b}\in\mathbb{R}^9 \}.
  \end{split}
\label{eqn_addsuboperdef}
\end{equation}

The addition and subtraction are actually the translation of the reachable set. 
\begin{myDef}
For a given terminal state $\boldsymbol{x}_T \sim \mathcal{N}(\boldsymbol{x}_T,\boldsymbol{\Sigma}_T)$ and a reachable set ${^{\boldsymbol{F}}_{\boldsymbol{G}}}\boldsymbol{\mathcal{P}}(\boldsymbol{x}_0,T)$, the cover rate of the $\boldsymbol{x}_0$ to $\boldsymbol{x}_T$ is the probabilities of $\boldsymbol{x}_T$ located in ${^{\boldsymbol{F}}_{\boldsymbol{G}}}\boldsymbol{\mathcal{P}}(\boldsymbol{x}_0,T)$, noted as ${^{\boldsymbol{F}}_{\boldsymbol{G}}}C(\boldsymbol{x}_0,T,\boldsymbol{x}_T)=P_r(\boldsymbol{x}_T \in {^{\boldsymbol{F}}_{\boldsymbol{G}}}\boldsymbol{\mathcal{P}}(\boldsymbol{x}_0,T))$.
\label{coverrateDef}
\end{myDef}

Where, $\boldsymbol{\Sigma}_T$ is a diagonal matrix. The cover rate of $\boldsymbol{x}_0$ to $\boldsymbol{x}_T$ represents the probabilities of $\boldsymbol{x}_T$ located in the $T$ reachable set of $\boldsymbol{x}_0$. 

\begin{myDef}
For a given reachable state $\boldsymbol{x}_0$, a Gaussian joint distribution $\boldsymbol{X} \sim \mathcal{N}(\boldsymbol{x}_{0},\boldsymbol{\Sigma}_{0})$, and the probability density function is ${^{\boldsymbol{\Sigma}_{0}}_{\boldsymbol{x}_{0}}}f(\boldsymbol{x})$. The $\boldsymbol{x}_0$ is the optimal dexterous waypoint relative to a terminal state $\boldsymbol{x}_T \sim \mathcal{N}(\boldsymbol{x}_T,\boldsymbol{\Sigma}_T)$, if $\boldsymbol{x}_0$ maximizes $\int {^{\boldsymbol{F}}_{\boldsymbol{G}}C(\boldsymbol{x},T,\boldsymbol{x}_T)}{^{\boldsymbol{\Sigma}_{0}}_{\boldsymbol{x}_{0}}}f(\boldsymbol{x})d\boldsymbol{x}$, noted as:
\begin{footnotesize}
\begin{equation}
  {^{\boldsymbol{F}}_{\boldsymbol{G}}}\boldsymbol{x}_{odw}(T,\boldsymbol{x}_T,\boldsymbol{\Sigma}_{0},\boldsymbol{\Sigma}_{T})=\mathop{\arg\max}\limits_{\boldsymbol{x}_{0}} \int  {^{\boldsymbol{F}}_{\boldsymbol{G}}C(\boldsymbol{x},T,\boldsymbol{x}_T)}{^{\boldsymbol{\Sigma}_{0}}_{\boldsymbol{x}_{0}}}f(\boldsymbol{x})d\boldsymbol{x}
\end{equation}
\end{footnotesize}
\label{optimalwaypointDef}
\end{myDef}

Where, $\boldsymbol{\Sigma}_0$ is a diagonal matrix which represents the states transition variance targeting to $\boldsymbol{x}_0$. According to the definition, an ODW's reachable set should cover $\boldsymbol{x}_T$ as much as possible. Meanwhile, the near states around ${^{\boldsymbol{F}}_{\boldsymbol{G}}\boldsymbol{x}_{odw}}(T,\boldsymbol{x}_T,\boldsymbol{\Sigma}_{0},\boldsymbol{\Sigma}_{T})$ should also possess higher cover rate to $\boldsymbol{x}_T$ such that the waypoint is robust to tracking errors. That is if a quadrotor flies towards ${^{\boldsymbol{F}}_{\boldsymbol{G}}\boldsymbol{x}_{odw}}(T,\boldsymbol{x}_T,\boldsymbol{\Sigma}_{0},\boldsymbol{\Sigma}_{T})$, it will possess highest possibility to plan feasible trajectories from  ${^{\boldsymbol{F}}_{\boldsymbol{G}}\boldsymbol{x}_{odw}}(T,\boldsymbol{x}_T,\boldsymbol{\Sigma}_{0},\boldsymbol{\Sigma}_{T})$ to $\boldsymbol{x}_T$.

To employ the ODW in real-time planning, we introduce two properties without considering constraints (geometrical and kinematic constraints) on states. 
\begin{myProperty} 
  \begin{equation}
  \begin{split}
  & {^{\boldsymbol{F}}_{\ast}}\boldsymbol{\mathcal{P}}(\boldsymbol{x}_0,T)\oplus \Delta\boldsymbol{x}_T={^{\boldsymbol{F}}_{\ast}}\boldsymbol{\mathcal{P}}(\boldsymbol{x}_0+\Delta\boldsymbol{x}_{0},T),\\
  & if: \Delta\boldsymbol{x}_{0}=e^{-\boldsymbol{A}T}\Delta\boldsymbol{x}_T \quad and \quad \boldsymbol{A}=\left[
    \begin{array}{ccc}
      \boldsymbol{0} & \boldsymbol{I}_3 & \boldsymbol{0}\\
      \boldsymbol{0} & \boldsymbol{0} & \boldsymbol{I}_3\\
      \boldsymbol{0} & \boldsymbol{0} & \boldsymbol{0}
    \end{array}
    \right]
  \end{split}
  \label{eqn_reachablesettransform}
  \end{equation}
\end{myProperty} 

Where, $\boldsymbol{0}$ and $\boldsymbol{I}$ are an all-zeros matrix and an identity matrix with appropriate sizes, respectively. This property demonstrates that the translation $\Delta\boldsymbol{x}_T$ of a reachable set can be transformed into the translation $\Delta\boldsymbol{x}_0$ of the start point $\boldsymbol{x}_0$.
\begin{myProof}
The state equations of the system is $\dot{\boldsymbol{x}}=\boldsymbol{A}\boldsymbol{x}+\boldsymbol{B}\boldsymbol{u},\boldsymbol{B} \in \mathbb{R}^{9\times3}, \boldsymbol{u}\in\mathbb{R}^{3} $. Select one element $\boldsymbol{x}_{Ti} \in {^{\boldsymbol{F}}_{\ast}}\boldsymbol{\mathcal{P}}(\boldsymbol{x}_0,T)$ and the corresponding control input is $\boldsymbol{u}_i(t)\in{^{\boldsymbol{F}}_{\ast}}\boldsymbol{\mathcal{U}}(\boldsymbol{x}_0,T,t),t \in [0,T]$. Then, 
\begin{equation}
\boldsymbol{x}_{Ti}=e^{\boldsymbol{A}T}\boldsymbol{x}_0+\int_0^Te^{\boldsymbol{A}(T-\tau)}B\boldsymbol{u}_i(\tau)d\tau
\label{eqn_oldterminalstateequation}
\end{equation}
and
\begin{equation}
  \begin{split}
& e^{\boldsymbol{A}T}(\boldsymbol{x}_0+\Delta\boldsymbol{x}_{0})+\int_0^Te^{\boldsymbol{A}(T-\tau)}B\boldsymbol{u}_i(\tau)d\tau \\
& =\boldsymbol{x}_{Ti}+e^{\boldsymbol{A}T}\Delta\boldsymbol{x}_{0}=\boldsymbol{x}_{Ti}+\Delta\boldsymbol{x}_T
\label{eqn_newterminalstateequation}
  \end{split}
\end{equation}
Combine (\ref{eqn_oldterminalstateequation})-(\ref{eqn_newterminalstateequation}), we get 
\begin{equation}
{^{\boldsymbol{F}}_{\ast}}\boldsymbol{\mathcal{P}}(\boldsymbol{x}_0+\Delta\boldsymbol{x}_{0},T)\supseteq {^{\boldsymbol{F}}_{\ast}}\boldsymbol{\mathcal{P}}(\boldsymbol{x}_0,T)\oplus\Delta\boldsymbol{x}_T. 
\label{eqn_forwardbelongproofed}
\end{equation}
Select one element $\boldsymbol{x}_{Ti}^\prime \in {^{\boldsymbol{F}}_{\ast}}\boldsymbol{\mathcal{P}}(\boldsymbol{x}_0+\Delta\boldsymbol{x}_{0},T)$ and corresponding input is $\boldsymbol{u}_i^\prime(t)\in{^{\boldsymbol{F}}_{\ast}}\boldsymbol{\mathcal{U}}(\boldsymbol{x}_0+\Delta\boldsymbol{x}_{0},T,t),t \in [0,T]$. Then,
\begin{footnotesize}
\begin{equation}
  \begin{split}
    & \boldsymbol{x}_{Ti}^\prime-\Delta\boldsymbol{x}_T=e^{\boldsymbol{A}T}(\boldsymbol{x}_0+\Delta\boldsymbol{x}_{0})+\int_0^Te^{\boldsymbol{A}(T-\tau)}B\boldsymbol{u}_i^\prime(\tau)d\tau \\
    & -e^{\boldsymbol{A}T}\Delta\boldsymbol{x}_{0}=e^{\boldsymbol{A}T}\boldsymbol{x}_0+\int_0^Te^{\boldsymbol{A}(T-\tau)}B\boldsymbol{u}_i^\prime(\tau)d\tau) \nonumber
    \label{eqn_reflectproof}
  \end{split}
\end{equation}
\end{footnotesize}
That means ${^{\boldsymbol{F}}_{\ast}}\boldsymbol{\mathcal{P}}(\boldsymbol{x}_0+\Delta\boldsymbol{x}_{0},T)\ominus\Delta\boldsymbol{x}_T\subseteq {^{\boldsymbol{F}}_{\ast}}\boldsymbol{\mathcal{P}}(\boldsymbol{x}_0,T)$. 
Further, we can get 
\begin{footnotesize}
\begin{equation}
  \begin{split}
   & {^{\boldsymbol{F}}_{\ast}}\boldsymbol{\mathcal{P}}(\boldsymbol{x}_0+\Delta\boldsymbol{x}_{0},T)\ominus\Delta\boldsymbol{x}_T\oplus \Delta\boldsymbol{x}_T \subseteq {^{\boldsymbol{F}}_{\ast}}\boldsymbol{\mathcal{P}}(\boldsymbol{x}_0,T) \oplus \Delta\boldsymbol{x}_T \\
   & \Rightarrow {^{\boldsymbol{F}}_{\ast}}\boldsymbol{\mathcal{P}}(\boldsymbol{x}_0+\Delta\boldsymbol{x}_{0},T)\subseteq {^{\boldsymbol{F}}_{\ast}}\boldsymbol{\mathcal{P}}(\boldsymbol{x}_0,T) \oplus \Delta\boldsymbol{x}_T
  \end{split} \label{eqn_reflectbelongproofed}
\end{equation}
\end{footnotesize}
Combine (\ref{eqn_reflectbelongproofed})-(\ref{eqn_forwardbelongproofed}), we get ${^{\boldsymbol{F}}_{\ast}}\boldsymbol{\mathcal{P}}(\boldsymbol{x}_0+\Delta\boldsymbol{x}_{0},T)= {^{\boldsymbol{F}}_{\ast}}\boldsymbol{\mathcal{P}}(\boldsymbol{x}_0,T) \oplus \Delta\boldsymbol{x}_T$ and ${^{\boldsymbol{F}}_{\ast}}\boldsymbol{\mathcal{U}}(\boldsymbol{x}_0,T,t)={^{\boldsymbol{F}}_{\ast}}\boldsymbol{\mathcal{U}}(\boldsymbol{x}_0+\Delta\boldsymbol{x}_{0},T,t)$.
\label{property_reachablesettranslation}
\end{myProof}

According to property (\ref{property_reachablesettranslation}), if $\Delta\boldsymbol{x}_{0}=e^{-\boldsymbol{A}T}\Delta\boldsymbol{x}_T$,we can get a remark as following:
\begin{myRemark}
\begin{equation}
  \begin{split}
    & {^{\boldsymbol{F}}_{\ast}}C(\boldsymbol{x}_0,T,\boldsymbol{x}_T)= P_r(\boldsymbol{x}_T \in{^{\boldsymbol{F}}_{\ast}}\boldsymbol{\mathcal{P}}(\boldsymbol{x}_0,T))\\
    & =P_r(\boldsymbol{x}_T+\Delta\boldsymbol{x}_T \in {^{\boldsymbol{F}}_{\ast}}\boldsymbol{\mathcal{P}}(\boldsymbol{x}_0,T)\oplus \Delta\boldsymbol{x}_T)\\
    & =P_r(\boldsymbol{x}_T+\Delta\boldsymbol{x}_T \in {^{\boldsymbol{F}}_{\ast}}\boldsymbol{\mathcal{P}}(\boldsymbol{x}_0+\Delta\boldsymbol{x}_0,T))\\
    & ={^{\boldsymbol{F}}_{\ast}}C(\boldsymbol{x}_0+\Delta\boldsymbol{x}_0,T,\boldsymbol{x}_T+\Delta\boldsymbol{x}_T)
   \end{split} \label{eqn_coverratetranform}
  \end{equation}
  \label{remark_coverratetranslation}
\end{myRemark}

That means the translated $\boldsymbol{x}_0$ by $\Delta\boldsymbol{x}_0$ has the same cover rate to the translated $\boldsymbol{x}_T$ by $\Delta\boldsymbol{x}_T$, if $\Delta\boldsymbol{x}_{0}=e^{-\boldsymbol{A}T}\Delta\boldsymbol{x}_T$. Especially, the two pairs of $\{\boldsymbol{x}_0,\boldsymbol{x}_T\}$ have same control input.

\begin{myProperty}
  For two given terminal states $\boldsymbol{x}_T \sim \mathcal{N}(\boldsymbol{x}_{T},\boldsymbol{\Sigma}_{T}), \boldsymbol{x}_T^\ast \sim \mathcal{N}(\boldsymbol{x}_{T}^\ast,\boldsymbol{\Sigma}_{T})$, the corresponding ODW ${^{\boldsymbol{F}}_{\ast}}\boldsymbol{x}_{odw}(T,\boldsymbol{x}_T,\boldsymbol{\Sigma}_{0},\boldsymbol{\Sigma}_{T})$ and $ {^{\boldsymbol{F}}_{\ast}}\boldsymbol{x}_{odw}(T,\boldsymbol{x}_T^\ast,\boldsymbol{\Sigma}_{0},\boldsymbol{\Sigma}_{T})$ have the following relationship:
  \begin{footnotesize}
  \begin{equation}
    \begin{split}
      & {^{\boldsymbol{F}}_{\ast}}\boldsymbol{x}_{odw}(T,\boldsymbol{x}_T,\boldsymbol{\Sigma}_{0},\boldsymbol{\Sigma}_{T})+\Delta\boldsymbol{x}_{odw}={^{\boldsymbol{F}}_{\ast}}\boldsymbol{x}_{odw}(T,\boldsymbol{x}_T^*,\boldsymbol{\Sigma}_{0},\boldsymbol{\Sigma}_{T})\\
      & \Delta\boldsymbol{x}_{odw}=e^{-\boldsymbol{A}T}(\boldsymbol{x}_T^*-\boldsymbol{x}_T)
     \end{split} \label{eqn_odwtranform}
    \end{equation}
  \end{footnotesize}
  \label{property_odwtranform}
\end{myProperty}

\begin{myProof}
  According to remark (\ref{remark_coverratetranslation}), we can get:
  \begin{footnotesize}
  \begin{equation}
    \begin{split}
    & {^{\boldsymbol{F}}_{\ast}}C(\boldsymbol{x}_0,T,\boldsymbol{x}_T)={^{\boldsymbol{F}}_{\ast}}C(\boldsymbol{x}_0+\Delta\boldsymbol{x}_{odw},T,\boldsymbol{x}_T+e^{\boldsymbol{A}T}\Delta\boldsymbol{x}_{odw})\\
    & ={^{\boldsymbol{F}}_{\ast}}C(\boldsymbol{x}_0+\Delta\boldsymbol{x}_{odw},T,\boldsymbol{x}_T^\ast)
    \end{split} \label{eqn_coverratepreproof}
  \end{equation}
  \end{footnotesize}
  Then, 
  \begin{footnotesize}
  \begin{equation}
    \begin{split}
      & \int {^{\boldsymbol{F}}_{\ast}}C(\boldsymbol{x},T,\boldsymbol{x}_T){^{\boldsymbol{\Sigma}_{0}}_{\boldsymbol{x}_{0}}}f(\boldsymbol{x})d\boldsymbol{x}\\
      & =\int {^{\boldsymbol{F}}_{\ast}}C(\boldsymbol{x},T,\boldsymbol{x}_T){^{\qquad \boldsymbol{\Sigma}_{0}}_{\boldsymbol{x}_{0}+\Delta\boldsymbol{x}_{odw}}}f(\boldsymbol{x}+\Delta\boldsymbol{x}_{odw})d\boldsymbol{x} \\
      & =\int {^{\boldsymbol{F}}_{\ast}}C(\boldsymbol{x}+\Delta\boldsymbol{x}_{odw},T,\boldsymbol{x}_T^\ast){^{\qquad \boldsymbol{\Sigma}_{0}}_{\boldsymbol{x}_{0}+\Delta\boldsymbol{x}_{odw}}}f(\boldsymbol{x}+\Delta\boldsymbol{x}_{odw})d({\boldsymbol{x}+\Delta\boldsymbol{x}_{odw}}) \nonumber
    \end{split}
    \end{equation}
  \end{footnotesize}
  Let $\boldsymbol{x}_{0}^\ast=\boldsymbol{x}_{0}+\Delta\boldsymbol{x}_{odw},\boldsymbol{x}^\ast=\boldsymbol{x}+\Delta\boldsymbol{x}_{odw}$, we can get:
  \begin{footnotesize}
    \begin{equation}
      \begin{split}
        & \int {^{\boldsymbol{F}}_{\ast}}C(\boldsymbol{x},T,\boldsymbol{x}_T){^{\boldsymbol{\Sigma}_{0}}_{\boldsymbol{x}_{0}}}f(\boldsymbol{x})d\boldsymbol{x}=\int  {^{\boldsymbol{F}}_{\ast}}C(\boldsymbol{x}^\ast,T,\boldsymbol{x}_T^\ast){^{\boldsymbol{\Sigma}_{0}}_{\boldsymbol{x}_{0}^\ast}}f(\boldsymbol{x}^\ast)d{\boldsymbol{x}^\ast}
      \end{split}\label{eqn_averagecoverequ}
      \end{equation}
    \end{footnotesize}
In (\ref{eqn_averagecoverequ}), $\boldsymbol{x}_{0}$ is the only variable. Therefore, if $\boldsymbol{x}_{0}={^{\boldsymbol{F}}_{\ast}}\boldsymbol{x}_{odw}(T,\boldsymbol{x}_T,\boldsymbol{\Sigma}_{0},\boldsymbol{\Sigma}_{T})$ can maximize the left part of (\ref{eqn_averagecoverequ}), then $\boldsymbol{x}_{0}^\ast={^{\boldsymbol{F}}_{\ast}}\boldsymbol{x}_{odw}(T,\boldsymbol{x}_T,\boldsymbol{\Sigma}_{0},\boldsymbol{\Sigma}_{T})+\Delta\boldsymbol{x}_{odw}$ can maximize the right part. That is 
\begin{footnotesize}
    \begin{equation}
    \begin{split}
      &{^{\boldsymbol{F}}_{\ast}}\boldsymbol{x}_{odw}(T,\boldsymbol{x}_T^\ast,\boldsymbol{\Sigma}_{0},\boldsymbol{\Sigma}_{T})\!=\!\mathop{\arg\max}\limits_{\boldsymbol{x}_{0}^\ast} \int  {^{\boldsymbol{F}}_{\ast}}C(\boldsymbol{x}^\ast,T,\boldsymbol{x}_T^\ast){^{\boldsymbol{\Sigma}_{0}}_{\boldsymbol{x}_{0}^\ast}}f(\boldsymbol{x}^\ast)d{\boldsymbol{x}^\ast}\\
      &{^{\boldsymbol{F}}_{\ast}}\boldsymbol{x}_{odw}(T,\boldsymbol{x}_T^\ast,\boldsymbol{\Sigma}_{0},\boldsymbol{\Sigma}_{T})\!=\!{^{\boldsymbol{F}}_{\ast}}\boldsymbol{x}_{odw}(T,\boldsymbol{x}_T,\boldsymbol{\Sigma}_{0},\boldsymbol{\Sigma}_{T})+\Delta\boldsymbol{x}_{odw}\nonumber
    \end{split} \label{eqn_odwtranslationproofed}
    \end{equation}  
  \end{footnotesize}
  Therefore, property (\ref{property_odwtranform}) is proofed.
\end{myProof}

According to property (\ref{property_odwtranform}), if we have a pair of $\{\boldsymbol{x}_T, {^{\boldsymbol{F}}_{\ast}}\boldsymbol{x}_{odw}(T,\boldsymbol{x}_T,\boldsymbol{\Sigma}_{0},\boldsymbol{\Sigma}_{T})\}$, the corresponding ODW ${^{\boldsymbol{F}}_{\ast}}\boldsymbol{x}_{odw}(T,\boldsymbol{x}_T^\ast,\boldsymbol{\Sigma}_{0},\boldsymbol{\Sigma}_{T})$ for a new $\boldsymbol{x}_T^\ast$ can be derived directly from (\ref{eqn_odwtranform}). The property is also called the transformation rule of ODWs for different $\boldsymbol{x}_T$. Especially, if the acceleration components of $\boldsymbol{x}_T$ and $\boldsymbol{x}_T^\ast$ are the same, we can get:
\begin{equation}
  \begin{split}
   & \boldsymbol{x}_T^\ast-\boldsymbol{x}_T=[\delta\boldsymbol{p}_T^\top,\delta\boldsymbol{v}_T^\top,\boldsymbol{0}^\top]^\top\\
   & \Delta\boldsymbol{x}_{odw}=[\delta\boldsymbol{p}_T^\top-T\delta\boldsymbol{v}_T^\top,\delta\boldsymbol{v}_T^\top,\boldsymbol{0}^\top]^\top
  \end{split} \label{eqn_odwtranslationexample}
\end{equation}

It should be noted that these properties only works without considering the constraints on states. They contain velocity and acceleration constraints induced by kinematic limitation and position constraints caused by geometric limitation. Therefore, the derived ${^{\boldsymbol{F}}_{\ast}}\boldsymbol{x}_{odw}^\ast$ might violate these constraints. 

In reality, the $\boldsymbol{a}_T$ is constant if the inclination of the target surfaces does not change according to (\ref{eqn_terminaldetermined}). In this situation, the feasibility of acceleration is not changed in the deduction of (\ref{eqn_odwtranform}) owing to the acceleration components of $\Delta\boldsymbol{x}_{odw}$ is zero in (\ref{eqn_odwtranslationexample}). Additionally, the allowed velocity is usually large enough for quadrotor if air drag is ignored or compensated by controllers. Hence, the velocity is always deemed to be feasible. Taking into account the geometrical constraints, the height direction ($\boldsymbol{Z}$ in Fig.\ref{initialproblemdescribe}) is constrained by the ground. Fortunately, the fluctuation of the surface in height is relative small for ground vehicles. That is the $\delta\boldsymbol{p}_T$ and $\delta\boldsymbol{v}_T$ in (\ref{eqn_odwtranslationexample}) will not significantly change the ${^{\boldsymbol{F}}_{\ast}}\boldsymbol{x}_{odw}(T,\boldsymbol{x}_T^\ast,\boldsymbol{\Sigma}_{0},\boldsymbol{\Sigma}_{T})$ in $\boldsymbol{Z}$ direction. Therefore, we can exploit property (\ref{property_odwtranform}) in $\boldsymbol{Z}$ direction in reality. Other directions $\boldsymbol{X},\boldsymbol{Y}$ have no geometrical conditions such that property (\ref{property_odwtranform}) can be used well. Therefore, the transformation rule is still available in practice.

\subsection{Seeking of Local Optimal Dexterous Waypoints}
An intuitive way to attain an ODW w.r.t a given $\boldsymbol{x}_T \sim \mathcal{N}(\boldsymbol{x}_T,\boldsymbol{\Sigma}_T)$ is to evaluate all  $\boldsymbol{x}_0$ in state space using definition (\ref{optimalwaypointDef}). However, two problems will emerge in this way. One is the state space is very large and most of the space is noneffective. The other is that $\boldsymbol{x}_0$ is not necessarily reachable even it possesses largest cover rate. 
\begin{algorithm}[!tb]
  \caption{Local Optimal Dexterous Waypoint }
  \label{alg_lodw}
  \LinesNumbered 
  \KwIn{
   $T,\boldsymbol{x}_T,\boldsymbol{Z}_s,N$\\
  }
  \KwOut{sequential LODWs $\boldsymbol{x}_{lodw}[N]$\\
  }
  \KwData{ $\boldsymbol{\Sigma}_{0},\boldsymbol{\Sigma}_{T}=diag(T^2\boldsymbol{\sigma}_v^\top+\boldsymbol{\sigma}_p^\top,\boldsymbol{\sigma}_v^\top,\boldsymbol{0}^\top)$\\
  }
  $\boldsymbol{x}_{lodw}[0]=\boldsymbol{x}_T, i=1$\\
  \While{$i \textless N$}
  {
      $\mathcal{X} \leftarrow$ \emph{\small \textbf{TrajectoryScan}}$(\boldsymbol{x}_T,\boldsymbol{Z}_s)$\\
      \For{$\boldsymbol{x}_j(t)$ in $\mathcal{X}$}
      {
        $\mathcal{F}_p[j]=\boldsymbol{x}_j(T_{fj}-T)$
      }
      $\boldsymbol{x}_{lodw}[i] =\mathop{\arg\max}\limits_{\boldsymbol{x}_{0}\in \mathcal{F}_p} \int  {^{\boldsymbol{F}}_{\boldsymbol{G}}C(\boldsymbol{x},T,\boldsymbol{x}_T)}{^{\boldsymbol{\Sigma}_{0}}_{\boldsymbol{x}_{0}}}f(\boldsymbol{x})d\boldsymbol{x}$\\
      $\boldsymbol{x}_T=\boldsymbol{x}_{lodw}[i]$\\
      $i=i+1$\\
  }
\end{algorithm}

\begin{figure}[!tb]
  \centering
  \includegraphics[width=3.0in]{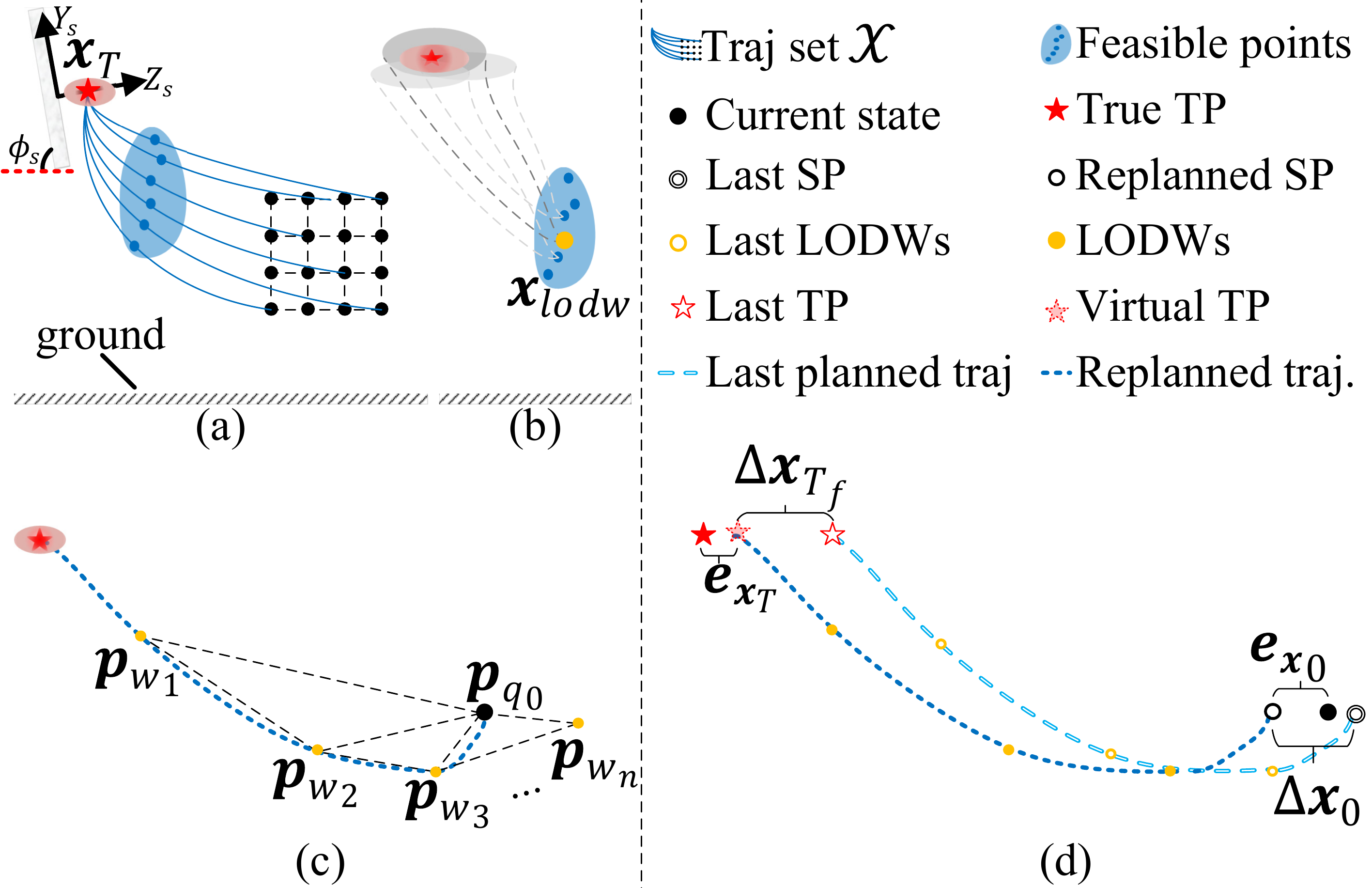}
  \caption{The flow to attain LODWs and method to exploit the LODWs in planning. (a-b) Three steps to attain an LODW. (c) The planning and replanning method using sequential LODWs. (d) The approximate complementary planner when there is no feasible trajectory.}
  \label{localodwmethod}
\end{figure}

To address these problems, we propose a method to obtain a local optimal dexterous waypoint (LODW) $\boldsymbol{x}_{lodw}(T,\boldsymbol{x}_T,\boldsymbol{\Sigma}_{0},\boldsymbol{\Sigma}_{T})$ for given $\boldsymbol{x}_T \sim \mathcal{N}(\boldsymbol{x}_T,\boldsymbol{\Sigma}_T)$ and $\boldsymbol{\Sigma}_0$. There are three steps shown in Fig.\ref{localodwmethod} (a-b) and Algorithm \ref{alg_lodw}. In the first step, we select a free position space on the side consistent with $\boldsymbol{Z}_{s_T}$. Then, the space is gridded.  The trajectories from the hover states of these discretized positions to $\boldsymbol{x}_T$ are planned (see Fig.\ref{localodwmethod}.a). These trajectories form a trajectory set $\mathcal{X}$ (Line 3, Algorithm \ref{alg_lodw}). In the second step, we select the point $\mathcal{F}_p[j]$ at $T_{fj}-T$ on every trajectory in $\mathcal{X}$. Where, $T_{fj}$ is the execution time of the $j$th trajectory. Obviously, these points are all reachable and effective. Therefore, the points form a feasible region where the LODWs would be chosen from (Line 4-6, Algorithm \ref{alg_lodw}). In the last step, we evaluate every point in the feasible region using definition (\ref{optimalwaypointDef}). The state that can maximizes $\int {^{\boldsymbol{F}}_{\boldsymbol{G}}C(\boldsymbol{x},T,\boldsymbol{x}_T)}{^{\boldsymbol{\Sigma}_{0}}_{\boldsymbol{x}_{0}}}f(\boldsymbol{x})d\boldsymbol{x}$ is selected as $\boldsymbol{x}_{lodw}(T,\boldsymbol{x}_T,\boldsymbol{\Sigma}_{0},\boldsymbol{\Sigma}_{T})$ (see Fig.\ref{localodwmethod}.b and Line 7 in Algorithm \ref{alg_lodw}). In particular, $\boldsymbol{x}_{lodw}[0]$ represents the target $\boldsymbol{x}_T$.

That is if the LODW is employed as a waypoint before target $\boldsymbol{x}_T$ by time $T$, the quadrotor will have larger probabilities to have feasible trajectory to reach the target in time $T$. Furthermore, we can get series of LODWs by updating $\boldsymbol{x}_T$ (see Fig.\ref{localodwmethod}.c and  Line 8 in Algorithm \ref{alg_lodw}). In this article, we assume the velocities of the surface are constant in every prediction horizon. The variances of the predicted position are induced by the predicted velocity variances $\boldsymbol{\sigma}_v$ and the detected position variances $\boldsymbol{\sigma}_p$. Herein, $\boldsymbol{\sigma}_v$ contains the variances induced by the detection noise and velocity fluctuation of the target. The variances of the predicted position would be enlarged if the prediction horizon increases.

\subsection{Trajectory Planning using Optimized Waypoints}
\begin{algorithm}[!tb]
  \caption{Planning and Replanning Method}
  \label{alg_replanning}
  \LinesNumbered 
  \KwIn{
   $\boldsymbol{s}(t)$, last solved trajectory $\boldsymbol{x}(t)_{ls}$, last solved execution time $T_{f_{ls}}$, 
   current state $\boldsymbol{x}_{q_0}$\\
  }
  \KwOut{replanned trajectory $\boldsymbol{x}(t)$\\
  }
  \KwData{ $T,\boldsymbol{x}_T,\boldsymbol{x}_{lodw}[N]$\\
  }

  \eIf{\emph{\textbf{FindTime}}$(\boldsymbol{s}(t),\boldsymbol{x}_{q_0},T_{f_{ls}})$==True}
  {
    $T_f=T_{f_{solved}},T_{f_{ls}}=T_f$\\
    $\boldsymbol{x}_T^\ast \leftarrow$ \emph{\small \textbf{GetTerminalPoint}}$(\boldsymbol{s}(T_f)), i=1$\\
    \While{$i \textless N$}
    {
      $\boldsymbol{x}_{lodw}^\ast[i]=e^{-\boldsymbol{A}iT}(\boldsymbol{x}_T^\ast-\boldsymbol{x}_T)+\boldsymbol{x}_{lodw}[i]$\\
      $i=i+1$\\
    }
    $\boldsymbol{x}_{lodw}^\ast[k] \leftarrow$ \emph{\small \textbf{GetForwardNearestLODW}}$(\boldsymbol{x}_T,\boldsymbol{x}_{lodw}^\ast[N],\boldsymbol{x}_{q_0}) $\\

    $\boldsymbol{x}(t) \leftarrow$ \emph{\small \textbf{PiecewisePlan}}$(\boldsymbol{x}_T^\ast,T_f,k,\boldsymbol{x}_{lodw}^\ast[N],\boldsymbol{x}_{q_0})$\\
    $\boldsymbol{x}(t)_{ls}=\boldsymbol{x}(t)$
  }
  {
    $T_f=T_{f_{ls}}-T_{f_{elps}}$\\
    $\boldsymbol{x}_T^\ast \leftarrow$ \emph{\small \textbf{GetTerminalPoint}}$(\boldsymbol{s}(T_f)) $\\
    $\check{\boldsymbol{x}}_T^\ast \leftarrow$ \emph{\small \textbf{GetVirtualTerminalPoint}}$(\boldsymbol{x}_T^\ast$,\\
    $\boldsymbol{x}(T_{f_{ls}})_{ls},T_f,
    \boldsymbol{x}(T_{f_{elps}})_{ls},\boldsymbol{x}_{q_0}) $\\
    $\boldsymbol{x}(t) =e^{-\boldsymbol{A}(T_f-t)}(\check{\boldsymbol{x}}_T^\ast-\boldsymbol{x}(T_{f_{ls}})_{ls})+$\\
    $\boldsymbol{x}(T_{f_{elps}}+t)_{ls},t \in [0,T_f]$\\
  }
\end{algorithm}

After getting a series of LODWs, a trajectory that passes these optimized waypoints should be planned. The planning and replanning method is shown in Algorithm \ref{alg_replanning}.  The algorithm is evoked when predicted $\boldsymbol{x}_T$ has changed or the quadrotor receives updated estimation about its current state. 

In Algorithm \ref{alg_replanning}, the function \textit{FindTime} is used to seek minimal feasible time for the trajectory. The time is also the predicted rendezvous time for perching. If minimal feasible time $T_{f_{solved}}$ is found, the new terminal point $\boldsymbol{x}_T^\ast$ can be obtained from (\ref{eqn_terminaldetermined}) by function \textit{GetTerminalPoint}. Then, the new LODWs $\boldsymbol{x}_{lodw}^\ast[N]$ can be derived using property (\ref{property_odwtranform}) (Line 4-7, Algorithm \ref{alg_replanning}).  For given $\boldsymbol{x}_{lodw}^\ast[N]$, the quadrotor needs to select the nearest point in its forward direction as the next waypoint. This is implemented in function \textit{GetForwardNearestLODW}. The process is shown in Fig.\ref{localodwmethod} (c). Where, $\boldsymbol{p}_{w_i},(i=1,2\dots n)$ denotes the position components of updated LODWs. $\boldsymbol{p}_{w_0}$ is the position components of $\boldsymbol{x}_T^\ast$. $\boldsymbol{p}_{q_0}$ is the components of current quadrotor's state. We traverse $\boldsymbol{p}_{w_i}$ from $i=1$ to $i=N$ and check the value of $\overrightarrow{\boldsymbol{p}_{q_0}\boldsymbol{p}_{w_i}}\cdot \overrightarrow{\boldsymbol{p}_{w_i}\boldsymbol{p}_{w_{i-1}}}$. Once the $\overrightarrow{\boldsymbol{p}_{q_0}\boldsymbol{p}_{w_i}}\cdot \overrightarrow{\boldsymbol{p}_{w_i}\boldsymbol{p}_{w_{i-1}}}\textless 0$, the waypoint $\boldsymbol{x}_{lodw}[i-1]$ is selected as the nearest point in quadrotor's forward direction. For brevity, we replace $i-1$ with $k$ and replace $\boldsymbol{x}_{lodw}^\ast[i-1]$ with $\boldsymbol{x}_{lodw}^\ast[k]$. After the forward nearest LODW $\boldsymbol{x}_{lodw}^\ast[k]$ is obtained, the trajectory time from $\boldsymbol{x}_{q_0}$ to $\boldsymbol{x}_{lodw}^\ast[k]$ is determined as $T_f-(k-1)T$. the allocated time for every segment between $\boldsymbol{x}_{lodw}^\ast[N]$ is known to be $T$. Then, the piecewise polynomial planner  proposed in \cite{richter2016polynomial,letizia2021novel} can be exploited to plan a trajectory by function \textit{PiecewisePlan}.

The function \textit{FindTime} is based on the multimodal search method proposed in our previous work \cite{liu2022hitchhiker}. For the sake of brevity, we won't go into this in this article. If no feasible time is found after the timeout, a complementary planner is provided to bring the target states up to date. The planned trajectory is transformed from the last solved trajectory $\boldsymbol{x}(t)_{ls}$ using property \ref{property_odwtranform} (Line 12-17, Algorithm \ref{alg_replanning}). The function \textit{GetVirtualTerminalPoint} are formulated as:
\begin{gather}
  \boldsymbol{x}_0^\ast=e^{-\boldsymbol{A}(T_{f_{ls}}-T_{f_{elps}})}(\check{\boldsymbol{x}}_T-\boldsymbol{x}(T_{f_{ls}})_{ls})+\boldsymbol{x}(T_{f_{elps}})_{ls} \label{eqn_virtualstartpoint}\\
  \boldsymbol{e}_{\boldsymbol{x}_T}=\check{\boldsymbol{x}}_T-\boldsymbol{x}_T^\ast,\boldsymbol{e}_{\boldsymbol{x}_0}=\boldsymbol{x}_0^\ast-\boldsymbol{x}_{q_0}\label{eqn_twoenderrors}\\
  \check{\boldsymbol{x}}_T^\ast=\mathop{\arg\max}\limits_{\check{\boldsymbol{x}}_T} \boldsymbol{e}_{\boldsymbol{x}_T}^\top\boldsymbol{\Lambda}_1\boldsymbol{e}_{\boldsymbol{x}_T}+\boldsymbol{e}_{\boldsymbol{x}_0}^\top\boldsymbol{\Lambda}_2\boldsymbol{e}_{\boldsymbol{x}_0}\label{eqn_virtualterminalpoint}
\end{gather}

Where, $T_{f_{ls}}$ is the execution time of $\boldsymbol{x}(t)_{ls}$. $T_{f_{elps}}$ is the elapsed time since $\boldsymbol{x}(t)_{ls}$ is solved. Therefore, $T_f=T_{f_{ls}}-T_{f_{elps}}$ is the execution time of the rest of $\boldsymbol{x}(t)_{ls}$. $\boldsymbol{x}(T_{f_{elps}})_{ls}$ and $\boldsymbol{x}(T_{f_{ls}})_{ls}$ are the start and terminal points of the rest part of $\boldsymbol{x}(t)_{ls}$, respectively. $\check{\boldsymbol{x}}_T$ is a virtual terminal point of the replanned trajectory and $\boldsymbol{x}_T^\ast$ is the updated terminal point at current time (see Fig.\ref{localodwmethod}). Exploiting property (\ref{property_odwtranform}) on the pair $\{\boldsymbol{x}(T_{f_{ls}})_{ls}, \boldsymbol{x}(T_{f_{elps}})_{ls}\}$, we can get the corresponding start point $\boldsymbol{x}_0^\ast$ for $\check{\boldsymbol{x}}_T$ according to (\ref{eqn_virtualstartpoint}). A trajectory from $\boldsymbol{x}_0^\ast$ to $\check{\boldsymbol{x}}_T$ can be planned with the same control input as $\boldsymbol{x}(t)_{ls},t\in [T_{f_{elps}},T_{f_{ls}}]$. If let $\check{\boldsymbol{x}}_T=\boldsymbol{x}_T^\ast$, a trajectory targeting to $\boldsymbol{x}_T^\ast$ can be planned. However, the corresponding start point $\boldsymbol{x}_0^\ast$ might be significantly different with the quadrotor's current state $\boldsymbol{x}_{q_0}$. That may cause large control input and tracking overshot. To trade off the terminal errors $\boldsymbol{e}_{\boldsymbol{x}_T}$ and start errors $\boldsymbol{e}_{\boldsymbol{x}_0}$, we construct a quadratic programming weighted by $\boldsymbol{\Lambda}_1=diag(\lambda_{11},\cdots,\lambda_{19})$ and $\boldsymbol{\Lambda}_2=diag(1-\lambda_{11},\cdots,1-\lambda_{19})$, formulated as (\ref{eqn_virtualterminalpoint}). Considering the larger prediction error of $\boldsymbol{x}_T^\ast$ when $T_f$ is large, we adopt time-varying weight based on sigmoid function, expressed as:
\begin{gather} 
\lambda_{1i}=1-1/(1+e^{-k_T(T_f-T_o)}).\label{eqn_lamdasigmoid}
\end{gather} 

Where, $k_T$ and $T_o$ are constant parameters. According to (\ref{eqn_lamdasigmoid}), $\lambda_{1i}$ tends to be zero if the time to arrive terminal point is too large. Then, the start point of the planned trajectory will be close to the quadrotor's current state. This avoids the oscillation of quadrotor caused by large prediction variance of $\boldsymbol{x}_T^\ast$. If $T_f$ is small, $\lambda_{1i}$ tends to be one and the virtual terminal point tends to be predicted terminal point $\boldsymbol{x}_T^\ast$.

 After the optimized virtual terminal point $\check{\boldsymbol{x}}_T^\ast$ is determined, every point of trajectory $\boldsymbol{x}(t)_{ls},t\in [T_{f_{elps}},T_{f_{ls}}]$ can be transformed using property (\ref{property_odwtranform}). Then, we can get the replanned trajectory as is shown in Line 16-17 in Algorithm \ref{alg_replanning}. However, there are always errors in terminal or start point in the planned trajectory. Therefore, we call it the approximate complementary planner.

\section{Thrust Optimization Controller}
In this section, we will propose a control strategy of quadrotor's thrust during attitude tracking when a quadrotor approaching a target surface. In this article, we assume there is no significant change in the posture and velocity direction of the surface. Based on the assumption, the quadrotor can fly in the symmetry sagittal plane during perching (see Fig.\ref{initialproblemdescribe} (a)). The simplified model of the quadrotor in sagittal plane are \cite{hu2019time,hehn2012performance}:
  \begin{equation}
    \begin{split}
    a_{q_y} \! =\!-\dfrac{f}{m}\sin{\phi},
    a_{q_z} \!=\!\dfrac{f}{m}\cos{\phi}-g,
    \ddot\phi \!=\!\dfrac{M}{J}
    \end{split} \label{eqn_dynamics}
  \end{equation}
  
Where $a_{q_y}$ and $a_{q_z}$ are the entries of $\boldsymbol{a}_q$ denoting the acceleration component of the quadrotor in $Y$ and $Z$ direction. $f$ is the total thrust. $M$ is the moment induced by thrust differential between the rear and front rotors. $J$ is the rotation inertia of quadrotor about its $X_q$ axis (see Fig.\ref{initialproblemdescribe} (a)). 

In practice, there are two stages in control strategy during a quadrotor approaching an inclined surface: position-velocity tracking stage  and attitude tracking stage. To distinguish the two stages, we define a time threshold $T_\epsilon \ge 0$. If $T_f\ge T_\epsilon$, the position-velocity tracking controller is employed to track the planned position and velocity profiles in feedback manner (blue dashed line in Fig.\ref{initialproblemdescribe}) \cite{thomas2016aggressive}. If $T_f\textless T_\epsilon$, the attitude tracking stage is activated to control the quadrotor's attitude to align with target surface (red dashed line in Fig.\ref{initialproblemdescribe}). During attitude tracking stage, positions and velocities are only controlled by planned $\boldsymbol{a}(t)$ without feedback. This would result in terminal position and velocity errors caused by internal and external disturbances.

Although the reference attitude angle $\phi$ are determined by planned $\boldsymbol{a}(t)$ to align with the surface during attitude tracking stage, the input $f$ is still adjustable without affecting the attitude command. By exploiting the thrust, we propose a thrust control strategy based on optimal control principle to minimize terminal position and velocity errors. In this controller, we decompose $f$ into $f_p$ and $f_o$. That is $f=f_p+f_o$. Here, $f_p$ is the planned thrust determined by planned trajectory $\boldsymbol{x}(t)$. $f_o$ is the component to be optimized to adjust the final thrust $f$.  To show the process of optimizing $f_o$, we construct new state variables as followings:
\begin{equation}
  \begin{split}
 &f_p\!=\!\begin{cases}
  m\sqrt{a_y(t)^2+(a_z(t)+g)^2} & t\in[0,T_{f}]\\
  m\sqrt{a_y(T_{f})^2+(a_z(T_{f})+g)^2} & t\in[T_{f}, +\infty]
  \end{cases} \label{eqn_1}\\
 & \mathscr{X}_1 =p_{q_y}+\int_0^t(\int_0^\tau{\dfrac{f_p}{m}\sin(\phi)}dt)d\tau\\ \nonumber
 & \mathscr{X}_2 =v_{q_y}+\int_0^t{\dfrac{f_p}{m}\sin(\phi)}dt\\ \nonumber
 & \mathscr{X}_3 =p_{q_z}-\int_0^t(\int_0^\tau{\dfrac{f_p}{m}\cos(\phi)}dt)d\tau+\dfrac{gt^2}{2}\\ \nonumber
 & \mathscr{X}_4 =v_{q_z}-\int_0^t{\dfrac{f_p}{m}\cos(\phi)}dt+gt \nonumber
\end{split}
\end{equation}
Let $\boldsymbol{\mathscr{X}}=[\mathscr{X}_1,\mathscr{X}_2,\mathscr{X}_3,\mathscr{X}_4]^\top$ be a new state vector. Then, we get following state equation according to (\ref{eqn_dynamics}):
\begin{equation}
  \begin{split}
 & \dot{\boldsymbol{\mathscr{X}}}\!=\!\boldsymbol{\mathscr{A}}\boldsymbol{\mathscr{X}}\!+\!\boldsymbol{\mathscr{B}}(t)f_o, \boldsymbol{\mathscr{B}}(t)\!=\![0,-\dfrac{\sin(\phi)}{m},0,\dfrac{\cos(\phi)}{m}]^\top\\
 & \boldsymbol{\mathscr{A}}=\left[
  \begin{array}{cc}
    \boldsymbol{e}_2^\top & 0 \\
    \boldsymbol{0} & \boldsymbol{e}_2
  \end{array}
  \right], \ \boldsymbol{e}_2=[0,1,0]^\top
\end{split}
\end{equation}

To get the optimal thrust compensation $f_o^\ast$, we construct followings optimal control problem:
\begin{equation}
  \begin{split}
  & \mathop{\min}\limits_{f_o} \ J_{ocp} =(\boldsymbol{\mathscr{X}}(T_\mathscr{F})-\boldsymbol{\mathscr{X}}_{T})^\top\boldsymbol{\Gamma}(\boldsymbol{\mathscr{X}}(T_\mathscr{F})-\boldsymbol{\mathscr{X}}_{T})\\ 
  &  s.t. \boldsymbol{\mathscr{X}}(0)\!=\![p_{{q_y}_0},v_{{q_y}_0},p_{{q_z}_0},v_{{q_z}_0}]^\top, f_o \in [f_{ol},f_{ou}]\\
  \end{split} \label{eqn_ocp}
\end{equation}

Where, $T_\mathscr{F}$ is the terminal time. $\boldsymbol{\mathscr{X}}(T_\mathscr{F})$ is the terminal state. $\boldsymbol{\mathscr{X}}_T$ is the expected terminal state which is determined by $\boldsymbol{s}(t), \boldsymbol{x}_{T}$ and $T_\mathscr{F}$. $\boldsymbol{\Gamma}=diag(\gamma_1,\gamma_2,\gamma_3,\gamma_4)$ is the weight diagonal matrix. $\boldsymbol{\mathscr{X}}(0)$ is the initial state vector which is consistent with quadrotor's current state. $f_{ol}$ and $f_{ou}$ are the lower and upper bound of $f_o$, respectively. The optimal control is to find a $f_o$ to minimize the terminal errors $J_{ocp}$. The corresponding Hamiltonian function is $H(\boldsymbol{\eta},\boldsymbol{\mathscr{X}},f_o)=\boldsymbol{\eta}^\top(\boldsymbol{\mathscr{A}}\boldsymbol{\mathscr{X}}\!+\!\boldsymbol{\mathscr{B}}(t)f_o)$, $\boldsymbol{\eta} \in \mathbb{R}^4$ is the costate variable vector. The optimal control $f_o^\ast$ can be obtained from:
\begin{equation}
  \begin{split}
  &\dot{\boldsymbol{\eta}}^\top=-\dfrac{\partial H}{\partial \boldsymbol{\mathscr{X}}}=-\boldsymbol{\eta}^\top\boldsymbol{\mathscr{A}},\ f_o^\ast= \mathop{\arg\min}\limits_{f_o}\boldsymbol{\eta}^\top\boldsymbol{\mathscr{B}}(t){f_o}\\
  &\Rightarrow f_o^\ast=\dfrac{f_{ou}+f_{ol}}{2}-\dfrac{f_{ou}-f_{ol}}{2}sign(\boldsymbol{\eta}^\top\boldsymbol{\mathscr{B}}(t)).  
  \end{split} \label{eqn_ocpclosedsolved}
\end{equation}

Apparently, the solution of $f_o^\ast$ is Bang-Bang style and the switching between $f_{ol}$ and $f_{ou}$ occurs at the zeros of $\boldsymbol{\eta}^\top\boldsymbol{\mathscr{B}}(t)$. However, the costate variables $\boldsymbol{\eta}$ are hard to be determined and $\boldsymbol{\mathscr{B}}(t)$ also contains nonlinear trigonometric function. That makes the closed solution impossible. To solve the problem, we select switching time of $f_o^\ast$ as the decision variables to minimize $J_{ocp}$, detailed as follows:
\begin{equation}
  \begin{split}
    & \mathop{\min}\limits_{t_1,\cdots,t_m}  J_{ocp} =(\boldsymbol{\mathscr{X}}(T_\mathscr{F})\!-\!\boldsymbol{\mathscr{X}}_{T})^\top\boldsymbol{\Gamma}(\boldsymbol{\mathscr{X}}(T_\mathscr{F})\!-\!\boldsymbol{\mathscr{X}}_{T})\\
    & \boldsymbol{\mathscr{X}}({T_\mathscr{F}})\!=\!e^{\boldsymbol{\mathscr{A}}T_\mathscr{F}}\boldsymbol{\mathscr{X}}(0)\!+\!\int_0^{T_\mathscr{F}}\!e^{\boldsymbol{\mathscr{A}}(T_\mathscr{F}-\tau)}\boldsymbol{\mathscr{B}}(\tau)f_o^\ast(\tau)d\tau\\
     & f_o^\ast(t,t_1\cdots t_m)\!=\!\begin{cases}
      f_{ol} & others\\
      f_{ou} & t_i \leq t\textless t_{i+1}, i \!\in \! \mathbb{O},i+1 \! \leq \! m\\
      f_{ou} & t_m \leq t\textless T_\mathscr{F}, m \!\in \! \mathbb{O}
      \end{cases}\\
     & t \in [0,T_\mathscr{F}],\ 0\leq t_1\leq t_2,\cdots,t_m\leq T_\mathscr{F},m\!=\!n+1.
  \end{split} \label{eqn_switchtime}
\end{equation}

Where, $t_i,i=1,\cdots,m$ are switching times to be optimized and $\mathbb{O}$ is the odd set. $m$ is the switching number which should not be less than the number of zeros of $\boldsymbol{\eta}^\top\boldsymbol{\mathscr{B}}(t)$. $n$ is the order of $\boldsymbol{\eta}^\top\boldsymbol{\mathscr{B}}(t)$, which is the largest number of zeros. In this article, we assume $f_o^\ast(0)=f_{ol}$. This assumption will not influence optimal solution when $m>n$ is satisfied. Because $t_1=0$ will change $f_o^\ast(0)$ to be $f_{ou}$ and the extra $t_i$ still have enough switching number to optimize the problem. In this way, the problem (\ref{eqn_ocp}) can be converted into a nonlinear programming (\ref{eqn_switchtime}) using switching times as decision variables.

During the terminal attitude tracking, $\sin$ and $\cos$ can be approximated as second order polynomial and the angle rate is usually approximately first order polynomial. Therefore, the maximal order of the entries of $\boldsymbol{\mathscr{B}}(t)$ is approximately $2$. Furthermore, according to (\ref{eqn_ocpclosedsolved}), the maximal order of $\eta_i$ is $1$. Consequently, $n$ is approximated as $3$ and $m$ is $4$ in this article.

Additionally, the specific expression of $\phi(t)$ is: 
$$
\phi(t)=\phi_0+\bar{\omega}t,T_\mathscr{F}=(\phi_s-\phi_0)/\bar{\omega},t\in [0,T_\mathscr{F}].
$$

Where, $\phi_0$ is the current roll angle of a quadrotor. $\bar{\omega}$ is the average planned angle rate of $\boldsymbol{x}(t),t\in[0,T_f]$ during the attitude tracking stage. $\phi_s$ is the inclination angle of target surface (see Fig.\ref{initialproblemdescribe}). $T_\mathscr{F}$ is determined to make the minimal  $J_{ocp}$ occur when quadrotor's angle is equal to $\phi_s$. It is worth noting that $T_\mathscr{F}$ might not be equal to $T_f$ owing to attitude tracking errors or delay in attitude loop. We assume the average angle rate $\bar{\omega}$ keep same for $t\in [T_f,T_\mathscr{F}]$.

\section{Experiments, Results and Discussion}
We conducted two groups of experiments to validate our proposed scheme. The first group of experiments is to validate the effectiveness of our proposed planner and thrust controller in simulation environments. The second group of experiments  is to verify the performance of a quadrotor perching on the rear window of a moving car outdoors.
\subsection{Comparison Experiments of Planner and Thrust Controller}
\begin{figure}[!tb]
  \centering
  \includegraphics[width=3.0in]{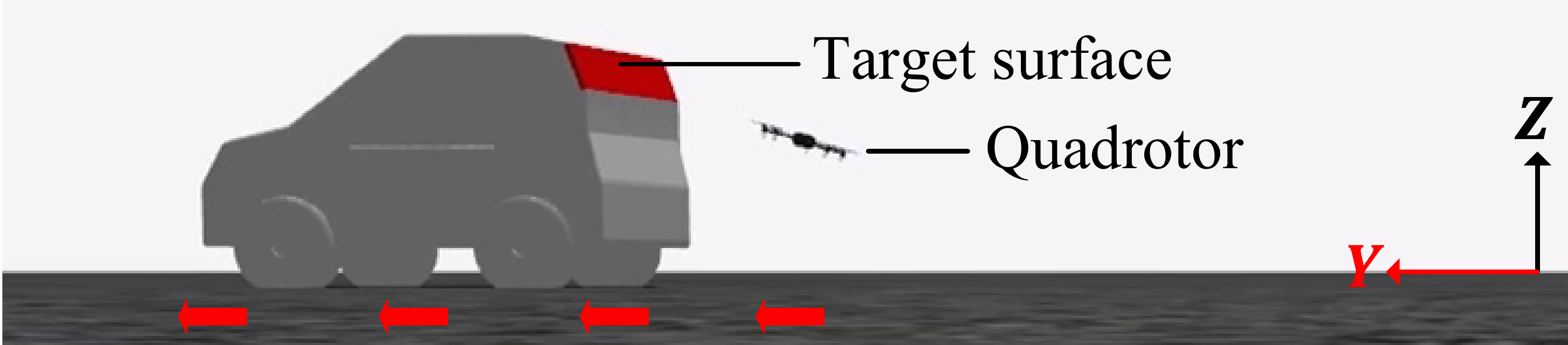}
  \caption{Diagram of the simulation experiments in Gazebo environment.}
  \label{simexp}
\end{figure}
\begin{figure}[!tb]
  \centering
  \includegraphics[width=3.0in]{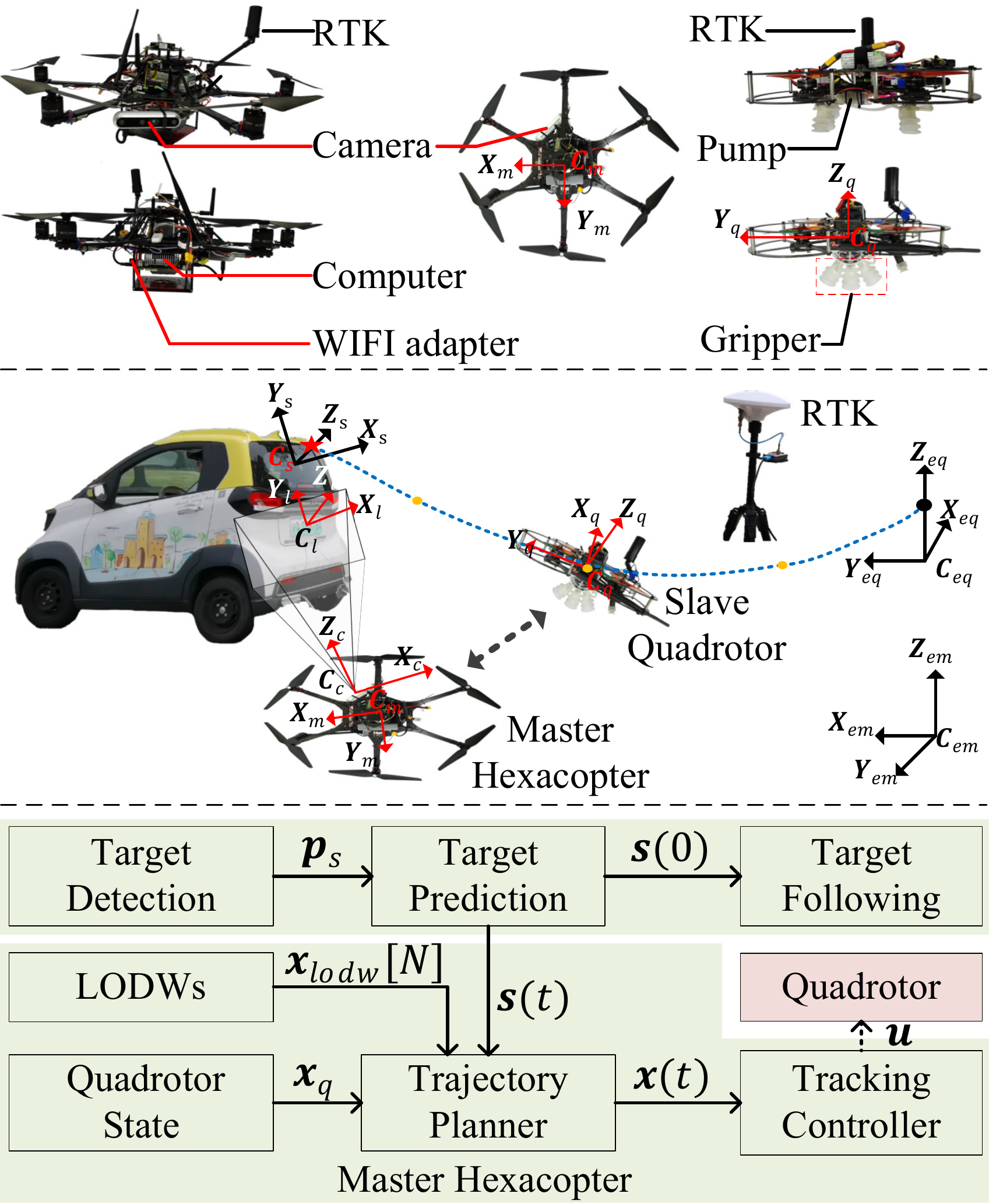}
  \caption{The heterogeneous cooperation detection and perch framework.}
  \label{heterogeneousframework}
\end{figure}
The comparison experiments are conducted in Gazebo simulation environments as is shown in Fig.\ref{simexp}. The frequency of control and state estimation is 30Hz. The thrust control method is implemented in C++11 with a general non-linear optimization solver NLopt\footnote{http://github.com/stevengj/nlopt}\cite{kraft1991tomp}. All programs are run on a ground computer (Intel-i7-6800k CPU @ 3.4GHz, Ubuntu-18.04). The target surface and the quadrotor move in $\boldsymbol{Y}-\boldsymbol{Z}$ plane. The target surface (red area in Fig.\ref{simexp}) is inclined with $\phi_s=70^\circ$. In these experiments, we focus on validating the performance of the trajectory planner and thrust optimization controller. Detected position of the target is simulated by adding random noises on the true position. The detection noises in $\boldsymbol{Y}$ and $\boldsymbol{Z}$ directions are set to the uniform distributions as ${dn}_y\sim{U}(-0.1,0.1)m,{dn}_z\sim{U}(-0.1,0.1)m$.  The car gradually accelerates to uniform linear motion (${v}_{s_y}=3m/s$) with a certain velocity fluctuations. The velocity fluctuations in $\boldsymbol{Y}$ and $\boldsymbol{Z}$ directions are set to the uniform  distributions as $v_{f_y}\sim{U}(-0.3,0.3)m/s,v_{f_z}\sim{U}(-0.1,0.1)m/s$. After the car's velocity exceeds a threshold, the quadrotor is commanded to execute perch maneuver starting from following mode. Once the collision or failure is checked, the quadrotor is commanded to hover. 

In the simulation experiments, we record the quadrotor's positions and velocities relative to the target surface when the quadrotor collides with it. If $p_{r_y}\in [p_{\tau_l},p_{\tau_u}],v_{r_\tau}\in[v_{\tau_l},v_{\tau_u}],v_{r_n}\in[v_{n_l},v_{n_u}],\phi_c \in[\phi_s+\phi_{e_l},\phi_s+\phi_{e_u}]$, we think the trial is successful. Otherwise, we consider the trial is failed. Where, $p_{r_y}$ is the relative position in $\boldsymbol{Y}_s$. $p_{\tau_l}=-0.15m,p_{\tau_u}=0.15m$ represent the size of the surface in $\boldsymbol{Y}_s$ direction. $v_{r_\tau},v_{r_n}$ are the relative velocity in tangential ($\boldsymbol{Y}_s$) and normal ($\boldsymbol{Z}_s$) directions, respectively. $v_{\tau_l}=-1m/s,v_{\tau_u}=2.2m/s$ and $v_{n_l}=-2m/s,v_{n_u}=-0.5m/s$ are the corresponding lower and upper bounds of the allowable range .  $\phi_c$ is the quadrotor's roll angle at impact. $\phi_s$ is the target inclination. $\phi_{e_l}=-0.31rad/s,\phi_{e_u}=0.31rad/s$ are the lower and  upper bounds of the allowable errors. The allowable range of velocities and angle errors are determined by gripper's adaptability. 

As a contrast, we use the existing two-end trajectory planner in \cite{thomas2016aggressive,liu2022hitchhiker} to compare with our planner. The attitude tracking controller in\cite{pi2021reinforcement,liu2022hitchhiker}  is used to compare with our thrust optimization controller. There are 10 trials in every session and the results are shown in Fig.\ref{plannercomp}, Fig.\ref{ocpcomp}, Table.\ref{table_plannercomparisonresults} and Table.\ref{table_controllercomparisonresults}.

\subsection{Outdoor Experiments of Perching on the Window of a Car}
The  experiment configurations are shown in Fig.\ref{heterogeneousframework}. The system contains a master hexacopter and a slave quadrotor. The states of the hexacopter and  quadrotor can be estimated from PX4 autopilot using RTK. The hexacopter is equipped with a Realsense D455 camera in its front right face. The camera is used to detect and locate the feature-rich place of moving car (license plate). The relative poses between the target surface and license plate, and between the hexacopter's world coordinate $C_{em}$ and the quadrotor's world coordinate  $C_{eq}$ are known in advance. Therefore, the position of the surface can be obtained in a unified coordinate system. Two adaptive grippers using self-sealing suction cups are mounted on the bottom of the quadrotor. The details of the gripper is described in our previous work\cite{liu2022hitchhiker}. 

The software structure is shown at the bottom of Fig.\ref{heterogeneousframework}. The green region contains target detection, target following, trajectory planner and tracking controller modules. All these modules are deployed on the companion computer (UP Xtreme Intel-i7-8565U CPU @ 4.6GHz, Ubuntu-18.04) mounted on the master hexacopter. The planner and tracking controller are running at $30 Hz$. The frequency of the state estimation of the drone and the target is $30Hz$. The command from the master hexacopter and the states of the quadrotor are transferred by WIFI modules. The inclination of the rear window of the car is approximate to $70^\circ$.

During the experiments, the master hexacopter hovers at a position where the license plate can be detected. Then, the hexacopter follows the car steadily to ensure the license plate located in the center of the camera's field of view. The slave quadrotor is also controlled by the master hexacopter to follow the car. Once the speed of the car execeeds the given threshold, the perch planning is activated. The quadrotor is commanded to fly towards the rear window of the car. Six trials are conducted with the different car's speed magnitudes ($0m/s, 1m/s, 2m/s, -1m/s$). The results are shown in Table.\ref{table_realtestresults}, Fig.\ref{planningprocess} and Fig.\ref{perchsnapshot}.

\subsection{Results}
\begin{figure}[!tb]
  \centering
  \includegraphics[width=3.0in]{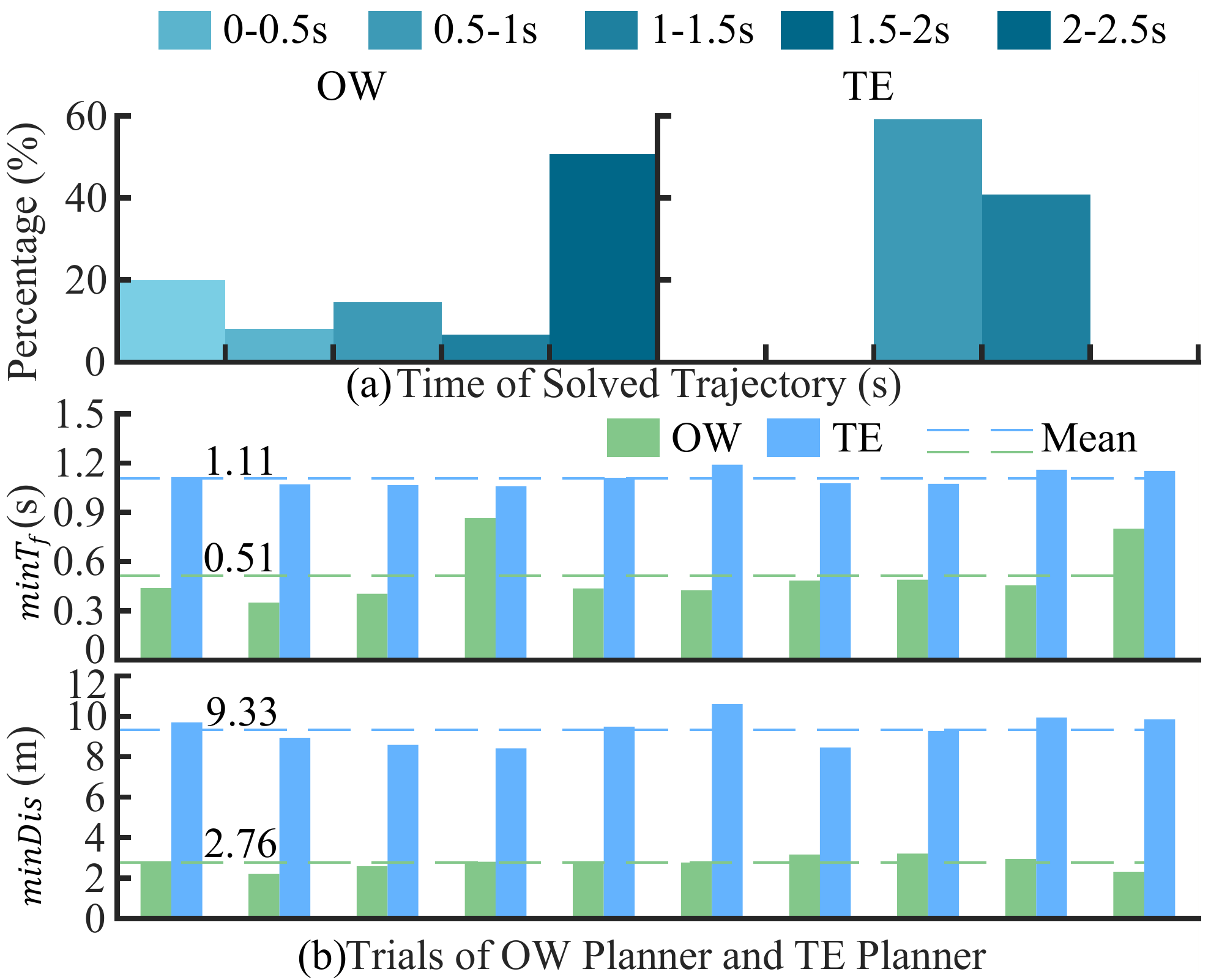}
  \caption{The results of comparison experiments for different planners. (a) The distribution of the duration of the solved trajectories. (b) The distribution of  $min T_f$ and $min {Dis}$ of  solved trajectories. The dashed lines in (b) are the means of the data.}
  \label{plannercomp}
\end{figure}
\begin{table}[!tb]
  \centering
  \renewcommand{\arraystretch}{1.0}
\begin{threeparttable}[b] 
  \caption{Average Results of Comparison Experiments of Planner}
  \label{table_plannercomparisonresults}
  \setlength{\tabcolsep}{1.0mm}
  {
    \begin{tabular}{c|cccccc}
      \toprule[1.2pt]
      &\begin{tabular}[c]{@{}c@{}}Success\\rate \end{tabular} & \begin{tabular}[c]{@{}c@{}}Avg\\ $min T_f$(s)\end{tabular} & \begin{tabular}[c]{@{}c@{}}Avg\\ $min Dis$(m)\end{tabular} & \begin{tabular}[c]{@{}c@{}}RMS\\ $p_{r_y}$(m)\end{tabular} & \begin{tabular}[c]{@{}c@{}}RMSE\\ $v_{r_y}$(m/s)\end{tabular} & \begin{tabular}[c]{@{}c@{}}RMSE\\ $v_{r_z}$(m/s)\end{tabular} \\ \midrule[1pt]
      OW & \textbf{5/10} & \textbf{0.51} & \textbf{2.76}  & \textbf{0.09}   & \textbf{1.05}  & \textbf{0.66}    \\ \cline{1-1}
      TE & 2/10 & 1.11 & 9.33 & 0.13  & 1.92  & 0.78 \\ 
      \bottomrule[1.2pt]
    \end{tabular}
  }
\end{threeparttable}
\end{table}

\begin{figure}[!tb]
  \centering
  \includegraphics[width=3.0in]{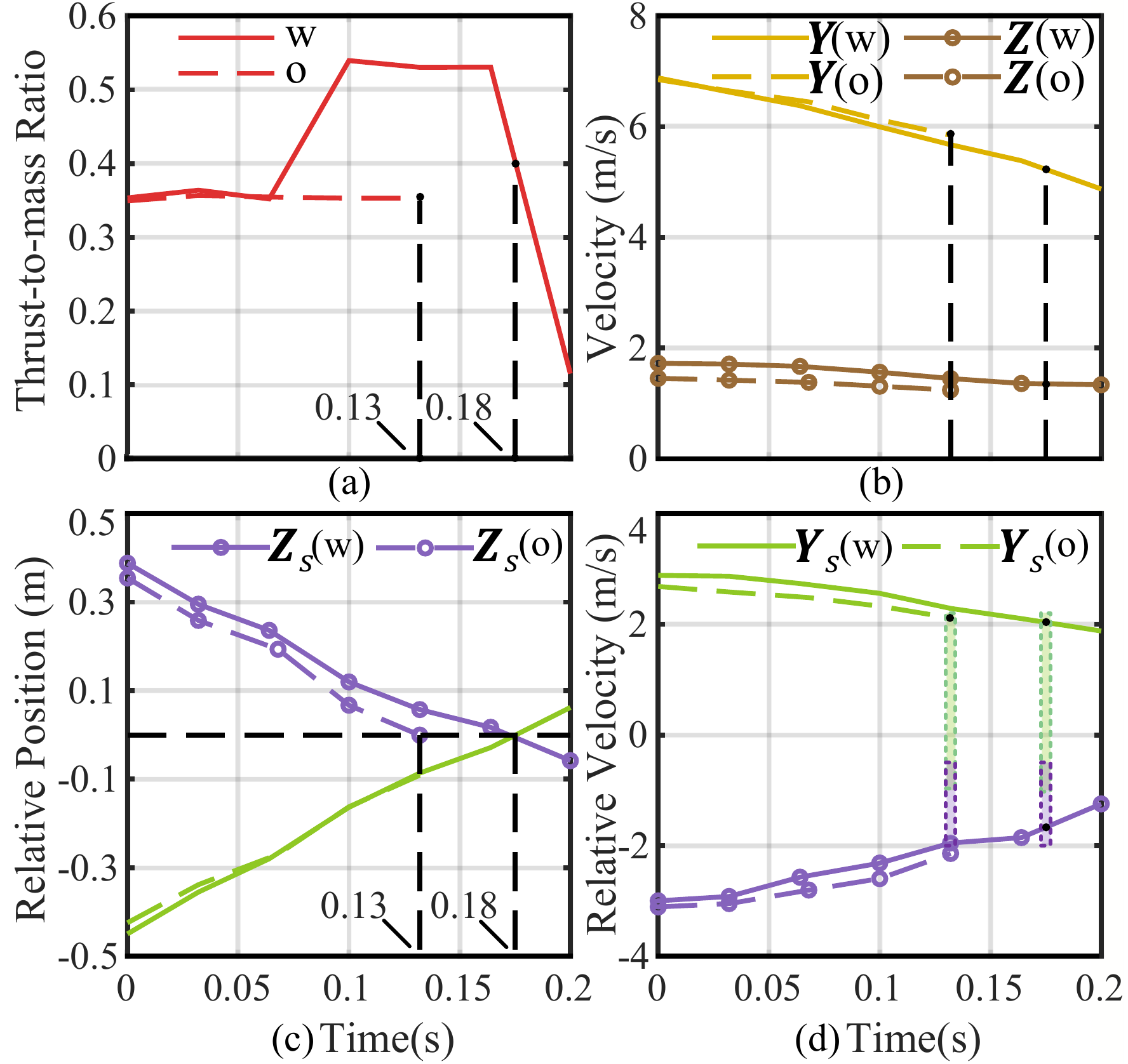}
  \caption{The profiles of two cases in comparison experiments for different controllers during attitude tracking stage. (a) The profiles of thrust-to-mass ratio. (b) The profiles of the quadrotor's velocities. (c) The profiles of relative positions. (d) The profiles of relative velocities.}
  \label{ocpcomp}
\end{figure}
\begin{table}[!tb]
  \centering
  \renewcommand{\arraystretch}{1.0}
\begin{threeparttable}[b] 
  \caption{Average Results of Comparison Experiments of Thrust Controller}
  \label{table_controllercomparisonresults}
  \setlength{\tabcolsep}{1.0mm}
  {
    \begin{tabular}{c|ccccc}
      \toprule[1.2pt]
      &\begin{tabular}[c]{@{}c@{}}Success\\rate \end{tabular} & \begin{tabular}[c]{@{}c@{}}Avg\\ $min T_f$(s)\end{tabular} & \begin{tabular}[c]{@{}c@{}}RMS\\ $p_{r_y}$(m)\end{tabular} & \begin{tabular}[c]{@{}c@{}}RMSE $v_{r_y}$\\(m/s)\end{tabular} & 
      \begin{tabular}[c]{@{}c@{}}RMSE $v_{r_z}$\\(m/s)\end{tabular}\\ \midrule[1pt]
      OW+TR & \textbf{7/10} & 0.66 & 0.1   & \textbf{1.00}  & \textbf{0.44}    \\ \cline{1-1}
      OW & 5/10 & 0.51 &  0.09   & 1.05  & 0.66   \\ 
      \bottomrule[1.2pt]
    \end{tabular}
  }
\end{threeparttable}
\end{table}
In Fig.\ref{plannercomp}, OW means the planner using optimized waypoints. TE means the conventional two-end planner. Fig.\ref{plannercomp} (a) demonstrates the distributions of the duration of the solved feasible trajectories. As we know, the execution time of the solved feasible trajectory is larger when the quadrotor is far from the target. And, the time is smaller when the quadrotor is getting closer to the target. We record the execution times of all solved feasible trajectories during perching. Then, we count the percentage of trajectories in different time intervals. The left and right parts are the results of OW and TE planner, respectively. We know that OW planner can solve feasible trajectories as the quadrotor approaches the target from a distance. However, TE planner can solve feasible trajectories only when the quadrotor is far from the target. Therefore, OW planner has a higher possibility of planning a feasible trajectory under the same uncertainties. We can use the time ($minT_f$) and distance ($minDis$) of the last planned feasible trajectory in each trial as an indicator. They are plotted in Fig.\ref{plannercomp} (b). We know that OW planner can plan feasible trajectories as the quadrotor gets closer to the target. 

Table.\ref{table_plannercomparisonresults} shows the average results of the planner comparison. RMS $p_{r_y}$ represent the  root-mean-square of tangential position. RMSE $v_{r_y}$ and RMSE $v_{r_z}$ represent the root-mean-square-error of the tangential and normal velocity relative to the centers of $[v_{\tau_l},v_{\tau_u}]$ and $[v_{n_l},v_{n_u}]$, respectively. The results demonstrate that the success rate of OW planner is higher. The average $minT_f$ and $minDis$ of OW planner are significantly smaller. The accuracies of position and velocity are also improved by using OW planner.

Fig.\ref{ocpcomp} shows the results of two cases in the attitude tracking stage with/without (w/o) thrust regulation (TR). The dashed lines show the profiles of the case without thrust regulation. The solid lines show the corresponding results with thrust optimization control. The green and purple boxes in Fig.\ref{ocpcomp} (d) represent the tolerable tangential and normal velocity of the gripper. In the two cases, the quadrotor's velocity in $\boldsymbol{Y}$ direction is large when the quadrotor enters the attitude tracking stage. For the case without thrust regulation, the quadrotor impacts on the target surface earlier due to the large velocity ($0.13s$ in Fig.\ref{ocpcomp}(c)). This results in excessive normal velocity at impact (Fig.\ref{ocpcomp}(d)) and failure. For the case with thrust optimization controller, the solved thrust by MPC will increase (Fig.\ref{ocpcomp} (a)). The increased thrust will reduce the velocity in $\boldsymbol{Y}_s$ directions more significantly. The moment of impact is delayed ($0.13s\textless0.18s$ in Fig.\ref{ocpcomp}(c)). The normal velocity at the moment of impact is also reduced to the tolerable range (Fig.\ref{ocpcomp}(d)). 

Table.\ref{table_controllercomparisonresults} shows the average results of the thrust controller. We can know that the success rate, tangential and normal velocities are improved by using thrust optimization controller.

\begin{table}[!tb]
  \centering
  \caption{Results of experiments for perching on the rear window of a car outdoors.}
  \label{table_realtestresults}
    \setlength{\tabcolsep}{1.8mm}{
    \begin{tabular}{c|cccccc}
      \toprule[1.2pt]
      Method & \multicolumn{4}{c|}{OW+TR} &OW+AT& TE+AT \\ \hline
      Trial No.& 1 & 2 & 3 & 4 & 5 & 6\\ \cline{1-1}
      Avg $v_{y_{car}}$(m/s) & 1.23 & 1.32 & 1.08 & \textbf{2.15} & -0.91 & 0 \\ \cline{1-1}
      Results & S & S & S & S & F & F \\ \cline{1-1}
      Avg min$T_f$(s) & 0.49 & \textbf{0.43} & 0.48 & 1.02  & 0.46 &\textbf{0.71} \\ \cline{1-1}
      Avg min Dis(m) & 1.54 & \textbf{1.51} & 2.06 & 1.41  & 1.8 &\textbf{3.99} \\ \cline{1-1}
      $p_{r_y}$(m) & -0.1 & -0.05 & 0.19 & 0.11  & 0.06 &  -0.06  \\ \cline{1-1}
      $v_{r_\tau}$(m/s) & -0.03 & 0.9 & -0.27 & 0.76  & 0.89 & \textbf{-1.09} \\ \cline{1-1}
      $v_{r_n}$(m/s) & -1.06 & -2.04 & -0.9 & -0.96  &{-2.28} &-0.95 \\ \cline{1-1}
      $\phi_e$(rad) & -0.04 & -0.3 & -0.3 & -0.24  & {-0.38}&-0.02    \\ \cline{1-1}
      ${dn}_y$(m) & 0.01 & 0.05 & 0.11 & \textbf{0.08}  & 0.06  &0.07 \\ \cline{1-1}
      ${dn}_z$(m) & 0.03 & 0.06 & 0.07 & 0.02  &0.03  &0.04\\ \cline{1-1}
      $v_{f_y}$(m/s) & 0 & 0.03 & 0.08 & \textbf{0.15}  & 0.06 &0.16 \\ \cline{1-1}
      $v_{f_z}$(m/s) & 0 & 0.02 & 0.05 & 0.01  & 0.01 & 0.01\\
      \bottomrule[1.2pt]
    \end{tabular}
  }
\end{table}
\begin{figure}[!tb]
  \centering
  \includegraphics[width=3.0in]{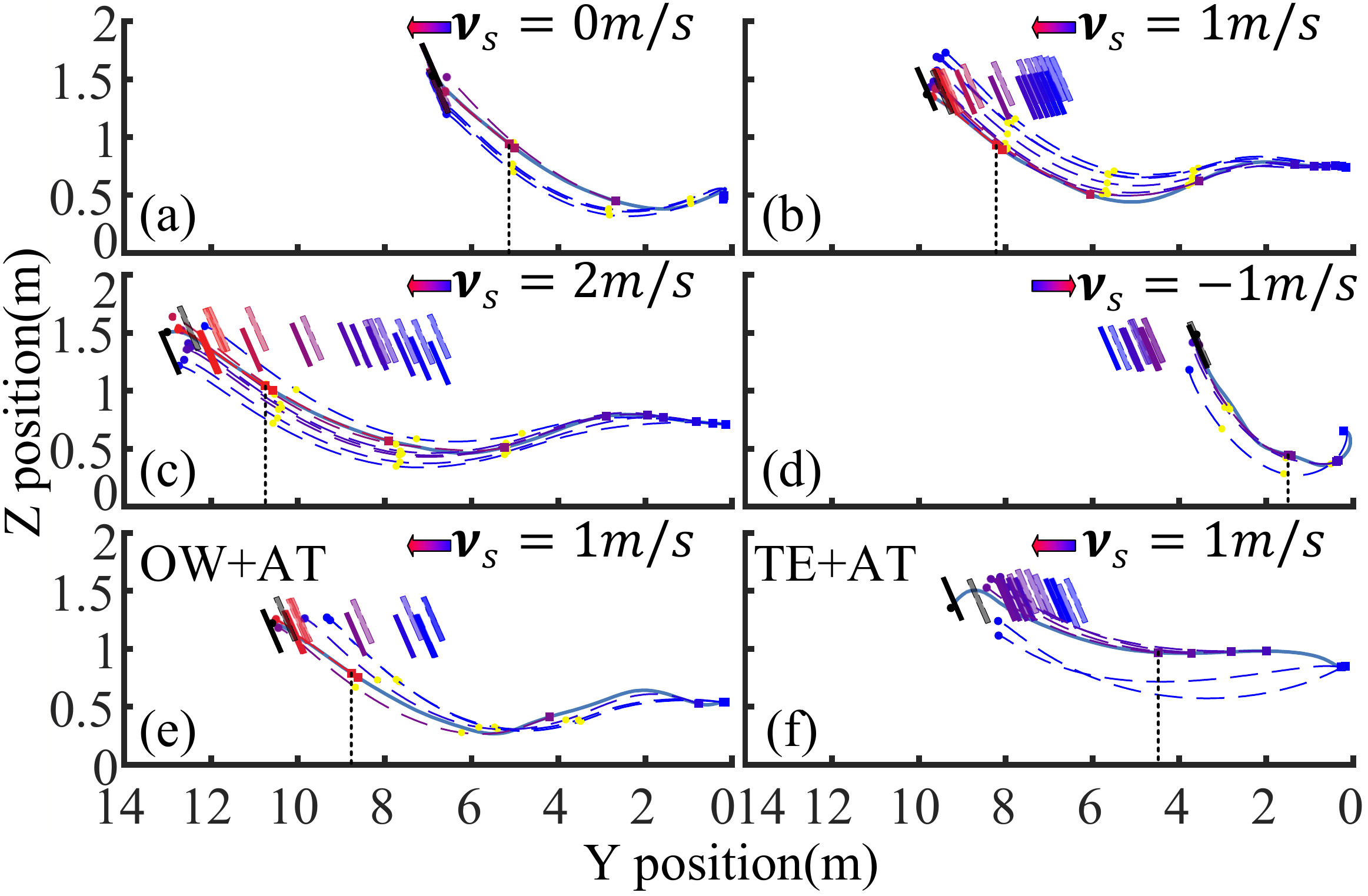}
  \caption{The trajectory planning process in outdoor perching on a car's window. (a-d) The process of perching on the rear window of car at $0m/s,1m/s,2m/s,-1m/s$ using our method. (e) The results of using OW+AT to perch on the car at $1m/s$. (f) The results of using TE+AT to perch on the car at $1m/s$.}
  \label{planningprocess}
\end{figure}
\begin{figure}[!tb]
  \centering
  \includegraphics[width=3.0in]{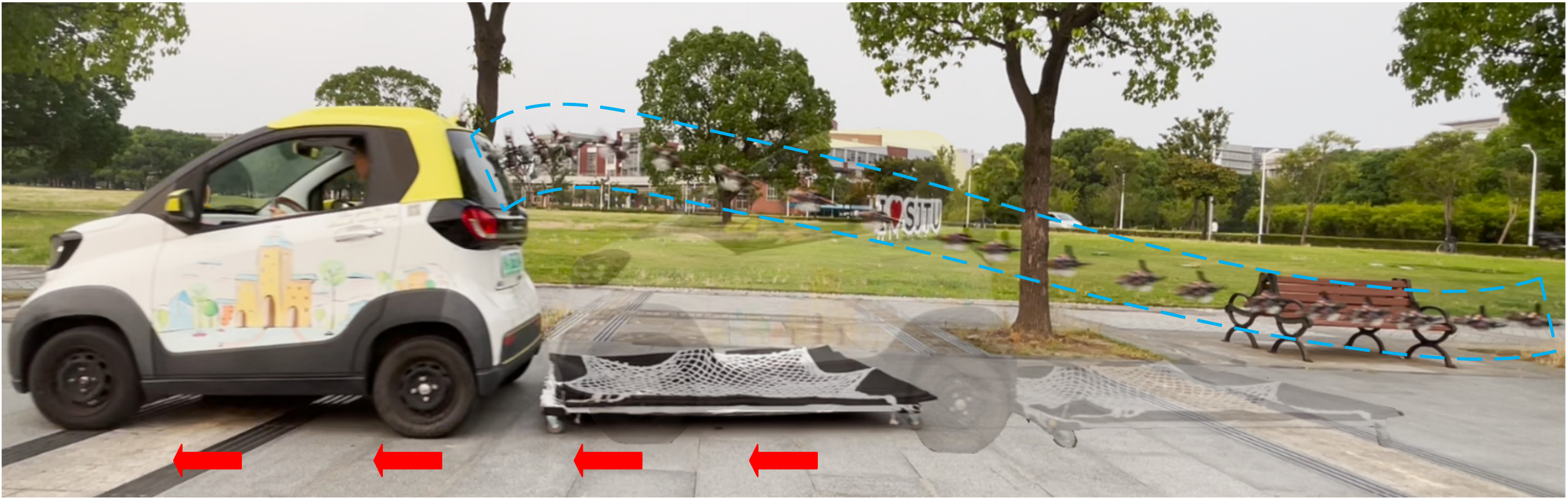}
  \caption{The snapshots of perching on the rear window of a moving car.}
  \label{perchsnapshot}
\end{figure}
The results of perching on real car's rear window is shown in Table.\ref{table_realtestresults}. The real-time planned trajectories with LODWs and actual trajectories during perching are presented in Fig.\ref{planningprocess}. Where, OW+TR means our proposed optimized waypoints planner and thrust regulation strategy. AT means the pure attitude tracking without thrust regulation. S and F represent successful and failed perch, respectively. $d_{n_y}$ and $d_{n_z}$ are the RMSE of detection noise based on the measurement of RTK modules on the car. $v_{f_y}$ and $v_{f_z}$ characterize the velocity fluctuations in reality. They are calculated by the RMSE of the car's velocity. $\phi_e$ is the impact angle error. In Fig.\ref{planningprocess}, the short colored rectangles show the positions of inclined window. Thereinto, the dark rectangles represent the true positions obtained from RTK. The light rectangle represent the detected positions obtained from camera. The dashed colored curves show the planned trajectory. The blue solid curves are the actual trajectories. The color from blue to red represents the passage of time. The dots and squares are the terminal points and start points of the solved trajectories, respectively. The dashed black lines demonstrate the position of the quadrotor when the last feasible trajectories are planned.  

From Table.\ref{table_realtestresults}, we can know that our proposed method enables the quadrotor to perch on the inclined window at different car's velocity. The perch can still be successful when the average velocity of the car is $2.15m/s$ and the detected noises in $\boldsymbol{Y}$ direction is $0.08m$ as well as the velocity fluctuation in $\boldsymbol{Y}$ direction is $0.15m/s$. Fig.\ref{planningprocess} demonstrates that LODWs (yellow dots) and trajectories can be updated with the movement of the car. Additionally, our planner can plan feasible trajectories when quadrotors approaches the target. The trajectory can be solved even the quadrotor is close to the targets. For TE planner, the trajectories can only be solved when the quadrotor is far from the target. The $minT_f$ and $minDis$ of OW is also significantly smaller
than that of TE ($0.43s\textless 0.71s, 1.51m\textless 3.99m$). Finally, the thrust regulation is also helpful for improving the terminal accuracy. The results of the real experiments are consistent with those of the simulation experiments. An image sequence of perching is shown in Fig.\ref{perchsnapshot} and the submitted video shows more details.

\subsection{Discussion}
In our proposed planner, the waypoints are optimized to adapt to uncertainties of predicted targets and tracking control. This increases the possibility that a feasible trajectory exists. Therefore, the OW planner can solve trajectories when the quadrotor approaches the target. Real-time update of trajectory is beneficial to improve the tracking and terminal accuracy. Furthermore, OW planner can update trajectories when the quadrotor is closer to the target. The shorter execution time of trajectories reduces the prediction time and improves the prediction accuracy. It is also helpful for enhancing terminal state accuracy.

The thrust optimization controller can regulate the thrust to minimize terminal errors. The controller solves a model predictive problem based on the quadrotor's current state. The terminal position and velocity are therefore closed loop. The immunity to disturbances and the terminal state accuracy are improved. 



\section{Conclusion and future work}
This article presents a trajectory planner and a thrust optimization controller to drive a quadrotor to perch on a moving inclined surface with outdoor uncertainties. In the trajectory planner, the waypoints are optimized by reachability analysis to adapt to the uncertainties in predicted targets and tracking control. In this way, trajectory passing the optimized waypoints possess large expected coverage for the distribution of predicted targets. The transformation rule of the optimized waypoints with respect to different targets is derived. Real-time planning based on optimized waypoints is developed. In the thrust optimization controller, the thrust is exploited to minimize the terminal errors when the attitude is commanded to be consistent with the target surface. An MPC problem is constructed and converted into a nonlinear programming using switching times as decision variables. The closed-loop control of terminal position and velocity is thus achieved in the attitude tracking stage. A heterogeneous cooperation framework containing a master hexacopter and a slave  quadrotor. With the framework, the consecutive and stable detection is achieved during perch.  The results of extensive simulation experiments demonstrate that it is easier to plan feasible trajectories using our planner under the same uncertainties. Our planner and control methods can improve the terminal states' accuracy. The success rate is increased by $50\%$ compare to the two-end planner without thrust regulation. The effectiveness and practicality of the methods are also verified by enabling a quadrotor to perch on the window of a moving car outdoors.

It should be pointed out that the transformation rule of LODWs is mainly used in the horizontal direction and slightly used in the height  direction by now. It is caused by constraints in kinodynamic and geometry. This motivates us to extend the library of LODW for target surfaces with different heights and inclinations in the future. Fitting and Interpolating skills will be employed to fetch local optimal waypoints from the library according to the detected target pose. Besides, the scheme using ego active vision to detect the target surfaces consecutively will also be explored.

\ifCLASSOPTIONcaptionsoff
  \newpage
\fi
\bibliographystyle{IEEEtran}
\bibliography{IEEEabrv,aggcupreference}

\begin{thebibliography}{10}
\providecommand{\url}[1]{#1}
\csname url@samestyle\endcsname
\providecommand{\newblock}{\relax}
\providecommand{\bibinfo}[2]{#2}
\providecommand{\BIBentrySTDinterwordspacing}{\spaceskip=0pt\relax}
\providecommand{\BIBentryALTinterwordstretchfactor}{4}
\providecommand{\BIBentryALTinterwordspacing}{\spaceskip=\fontdimen2\font plus
\BIBentryALTinterwordstretchfactor\fontdimen3\font minus
  \fontdimen4\font\relax}
\providecommand{\BIBforeignlanguage}[2]{{%
\expandafter\ifx\csname l@#1\endcsname\relax
\typeout{** WARNING: IEEEtran.bst: No hyphenation pattern has been}%
\typeout{** loaded for the language `#1'. Using the pattern for}%
\typeout{** the default language instead.}%
\else
\language=\csname l@#1\endcsname
\fi
#2}}
\providecommand{\BIBdecl}{\relax}
\BIBdecl

\bibitem{ji2022real}
J.~Ji, T.~Yang, C.~Xu, and F.~Gao, ``Real-time trajectory planning for aerial
  perching,'' \emph{arXiv preprint arXiv:2203.01061}, 2022.

\bibitem{zhang2020fast}
G.~Zhang, H.~Kuang, and X.~Liu, ``Fast trajectory optimization for quadrotor
  landing on a moving platform,'' in \emph{2020 International Conference on
  Unmanned Aircraft Systems (ICUAS)}.\hskip 1em plus 0.5em minus 0.4em\relax
  IEEE, 2020, pp. 238--246.

\bibitem{vlantis2015quadrotor}
P.~Vlantis, P.~Marantos, C.~P. Bechlioulis, and K.~J. Kyriakopoulos,
  ``Quadrotor landing on an inclined platform of a moving ground vehicle,'' in
  \emph{2015 IEEE International Conference on Robotics and Automation
  (ICRA)}.\hskip 1em plus 0.5em minus 0.4em\relax IEEE, 2015, pp. 2202--2207.

\bibitem{choudhury2021efficient}
S.~Choudhury, K.~Solovey, M.~J. Kochenderfer, and M.~Pavone, ``Efficient
  large-scale multi-drone delivery using transit networks,'' \emph{Journal of
  Artificial Intelligence Research}, vol.~70, pp. 757--788, 2021.

\bibitem{hsiao2022novel}
H.~Hsiao, F.~Wu, J.~Sun, and J.~Zhao, ``A novel passive mechanism for flying
  robots to perch onto surfaces,'' in \emph{2022 International Conference on
  Robotics and Automation (ICRA)}.\hskip 1em plus 0.5em minus 0.4em\relax IEEE,
  2022, pp. 1183--1189.

\bibitem{lasla2019exploiting}
N.~Lasla, H.~Ghazzai, H.~Menouar, and Y.~Massoud, ``Exploiting land transport
  to improve the uav's performances for longer mission coverage in smart
  cities,'' in \emph{2019 IEEE 89th Vehicular Technology Conference
  (VTC2019-Spring)}.\hskip 1em plus 0.5em minus 0.4em\relax IEEE, 2019, pp.
  1--7.

\bibitem{baca2019autonomous}
T.~Baca, P.~Stepan, V.~Spurny, D.~Hert, R.~Penicka, M.~Saska, J.~Thomas,
  G.~Loianno, and V.~Kumar, ``Autonomous landing on a moving vehicle with an
  unmanned aerial vehicle,'' \emph{Journal of Field Robotics}, vol.~36, no.~5,
  pp. 874--891, 2019.

\bibitem{dadkhah2012survey}
N.~Dadkhah and B.~Mettler, ``Survey of motion planning literature in the
  presence of uncertainty: Considerations for uav guidance,'' \emph{Journal of
  Intelligent \& Robotic Systems}, vol.~65, no.~1, pp. 233--246, 2012.

\bibitem{brault2021robust}
P.~Brault, Q.~Delamare, and P.~R. Giordano, ``Robust trajectory planning with
  parametric uncertainties,'' in \emph{2021 IEEE International Conference on
  Robotics and Automation (ICRA)}.\hskip 1em plus 0.5em minus 0.4em\relax IEEE,
  2021, pp. 11\,095--11\,101.

\bibitem{vermaelen2020survey}
J.~Vermaelen, H.~T. Dinh, and T.~Holvoet, ``A survey on probabilistic planning
  and temporal scheduling with safety guarantees,'' in \emph{ICAPS Workshop on
  Planning and Robotics}.\hskip 1em plus 0.5em minus 0.4em\relax ICAPS Workshop
  on Planning and Robotics, 2020.

\bibitem{tao2021path}
X.~Tao, N.~Lang, H.~Li, and D.~Xu, ``Path planning in uncertain environment
  with moving obstacles using warm start cross entropy,'' \emph{IEEE/ASME
  Transactions on Mechatronics}, 2021.

\bibitem{bey2021handling}
H.~Bey, M.~Sackmann, A.~Lange, and J.~Thielecke, ``Handling prediction model
  errors in planning for automated driving using pomdps,'' in \emph{2021 IEEE
  International Intelligent Transportation Systems Conference (ITSC)}.\hskip
  1em plus 0.5em minus 0.4em\relax IEEE, 2021, pp. 439--446.

\bibitem{ding2021epsilon}
W.~Ding, L.~Zhang, J.~Chen, and S.~Shen, ``Epsilon: An efficient planning
  system for automated vehicles in highly interactive environments,''
  \emph{IEEE Transactions on Robotics}, 2021.

\bibitem{zhu2019chance}
H.~Zhu and J.~Alonso-Mora, ``Chance-constrained collision avoidance for mavs in
  dynamic environments,'' \emph{IEEE Robotics and Automation Letters}, vol.~4,
  no.~2, pp. 776--783, 2019.

\bibitem{xu2021dpmpc}
Z.~Xu, D.~Deng, Y.~Dong, and K.~Shimada, ``Dpmpc-planner: A real-time uav
  trajectory planning framework for complex static environments with dynamic
  obstacles,'' \emph{arXiv preprint arXiv:2109.07024}, 2021.

\bibitem{wang2020non}
A.~Wang, A.~Jasour, and B.~C. Williams, ``Non-gaussian chance-constrained
  trajectory planning for autonomous vehicles under agent uncertainty,''
  \emph{IEEE Robotics and Automation Letters}, vol.~5, no.~4, pp. 6041--6048,
  2020.

\bibitem{zhang2020trajectory}
X.~Zhang, J.~Ma, Z.~Cheng, S.~Huang, S.~S. Ge, and T.~H. Lee, ``Trajectory
  generation by chance-constrained nonlinear mpc with probabilistic
  prediction,'' \emph{IEEE Transactions on Cybernetics}, vol.~51, no.~7, pp.
  3616--3629, 2020.

\bibitem{gueguen2009safety}
H.~Gu{\'e}guen, M.-A. Lefebvre, J.~Zaytoon, and O.~Nasri, ``Safety verification
  and reachability analysis for hybrid systems,'' \emph{Annual Reviews in
  Control}, vol.~33, no.~1, pp. 25--36, 2009.

\bibitem{kousik2019safe}
S.~Kousik, P.~Holmes, and R.~Vasudevan, ``Safe, aggressive quadrotor flight via
  reachability-based trajectory design,'' in \emph{Dynamic Systems and Control
  Conference}, vol. 59162.\hskip 1em plus 0.5em minus 0.4em\relax American
  Society of Mechanical Engineers, 2019, p. V003T19A010.

\bibitem{manzinger2020using}
S.~Manzinger, C.~Pek, and M.~Althoff, ``Using reachable sets for trajectory
  planning of automated vehicles,'' \emph{IEEE Transactions on Intelligent
  Vehicles}, vol.~6, no.~2, pp. 232--248, 2020.

\bibitem{lakhal2022safe}
N.~M.~B. Lakhal, L.~Adouane, O.~Nasri, and J.~B.~H. Slama, ``Safe and adaptive
  autonomous navigation under uncertainty based on sequential waypoints and
  reachability analysis,'' \emph{Robotics and Autonomous Systems}, vol. 152, p.
  104065, 2022.

\bibitem{yang2021learning}
R.~Yang, L.~Zheng, J.~Pan, and H.~Cheng, ``Learning-based predictive path
  following control for nonlinear systems under uncertain disturbances,''
  \emph{IEEE Robotics and Automation Letters}, vol.~6, no.~2, pp. 2854--2861,
  2021.

\bibitem{tran2019adaptive}
V.~P. Tran, F.~Santoso, M.~A. Garratt, and I.~R. Petersen, ``Adaptive
  second-order strictly negative imaginary controllers based on the interval
  type-2 fuzzy self-tuning systems for a hovering quadrotor with
  uncertainties,'' \emph{IEEE/ASME Transactions on Mechatronics}, vol.~25,
  no.~1, pp. 11--20, 2019.

\bibitem{wang2022integrated}
B.~Wang, Y.~Zhang, and W.~Zhang, ``Integrated path planning and trajectory
  tracking control for quadrotor uavs with obstacle avoidance in the presence
  of environmental and systematic uncertainties: Theory and experiment,''
  \emph{Aerospace Science and Technology}, vol. 120, p. 107277, 2022.

\bibitem{lu2020uncertainty}
Q.~Lu, B.~Ren, and S.~Parameswaran, ``Uncertainty and disturbance
  estimator-based global trajectory tracking control for a quadrotor,''
  \emph{IEEE/ASME Transactions on Mechatronics}, vol.~25, no.~3, pp.
  1519--1530, 2020.

\bibitem{hu2019time}
B.~Hu and S.~Mishra, ``Time-optimal trajectory generation for landing a
  quadrotor onto a moving platform,'' \emph{IEEE/ASME Transactions on
  Mechatronics}, vol.~24, no.~2, pp. 585--596, 2019.

\bibitem{sun2021comparative}
S.~Sun, A.~Romero, P.~Foehn, E.~Kaufmann, and D.~Scaramuzza, ``A comparative
  study of nonlinear mpc and differential-flatness-based control for quadrotor
  agile flight,'' \emph{arXiv preprint arXiv:2109.01365}, 2021.

\bibitem{pi2021reinforcement}
C.-H. Pi, K.-C. Hu, Y.-T. Huang, and S.~Cheng, ``Reinforcement learning
  trajectory generation and control for aggressive perching on vertical walls
  with quadrotors,'' \emph{arXiv preprint arXiv:2103.03011}, 2021.

\bibitem{liu2022hitchhiker}
S.~Liu, W.~Dong, Z.~Wang, and X.~Sheng, ``Hitchhiker: A quadrotor aggressively
  perching on a moving inclined surface using compliant suction cup gripper,''
  \emph{arXiv preprint arXiv:2203.02304}, 2022.

\bibitem{guo2020image}
D.~Guo and K.~K. Leang, ``Image-based estimation, planning, and control for
  high-speed flying through multiple openings,'' \emph{The International
  Journal of Robotics Research}, vol.~39, no.~9, pp. 1122--1137, 2020.

\bibitem{van2011lqg}
J.~Van Den~Berg, P.~Abbeel, and K.~Goldberg, ``Lqg-mp: Optimized path planning
  for robots with motion uncertainty and imperfect state information,''
  \emph{The International Journal of Robotics Research}, vol.~30, no.~7, pp.
  895--913, 2011.

\bibitem{richter2016polynomial}
C.~Richter, A.~Bry, and N.~Roy, ``Polynomial trajectory planning for aggressive
  quadrotor flight in dense indoor environments,'' in \emph{Robotics
  research}.\hskip 1em plus 0.5em minus 0.4em\relax Springer, 2016, pp.
  649--666.

\bibitem{letizia2021novel}
N.~A. Letizia, B.~Salamat, and A.~M. Tonello, ``A novel recursive smooth
  trajectory generation method for unmanned vehicles,'' \emph{IEEE Transactions
  on Robotics}, vol.~37, no.~5, pp. 1792--1805, 2021.

\bibitem{hehn2012performance}
M.~Hehn, R.~Ritz, and R.~D’Andrea, ``Performance benchmarking of quadrotor
  systems using time-optimal control,'' \emph{Autonomous Robots}, vol.~33,
  no.~1, pp. 69--88, 2012.

\bibitem{thomas2016aggressive}
J.~Thomas, M.~Pope, G.~Loianno, E.~W. Hawkes, M.~A. Estrada, H.~Jiang, M.~R.
  Cutkosky, and V.~Kumar, ``Aggressive flight with quadrotors for perching on
  inclined surfaces,'' \emph{Journal of Mechanisms and Robotics}, vol.~8,
  no.~5, p. 051007, 2016.

\bibitem{kraft1991tomp}
D.~Kraft, \emph{TOMP: FORTRAN modules for optimal control calculations}.\hskip
  1em plus 0.5em minus 0.4em\relax VDI-Verlag, 1991.

\end{thebibliography}

%

\vspace{-50 pt} 
\begin{IEEEbiography}[{\includegraphics[width=1in,height=1.2in,clip]{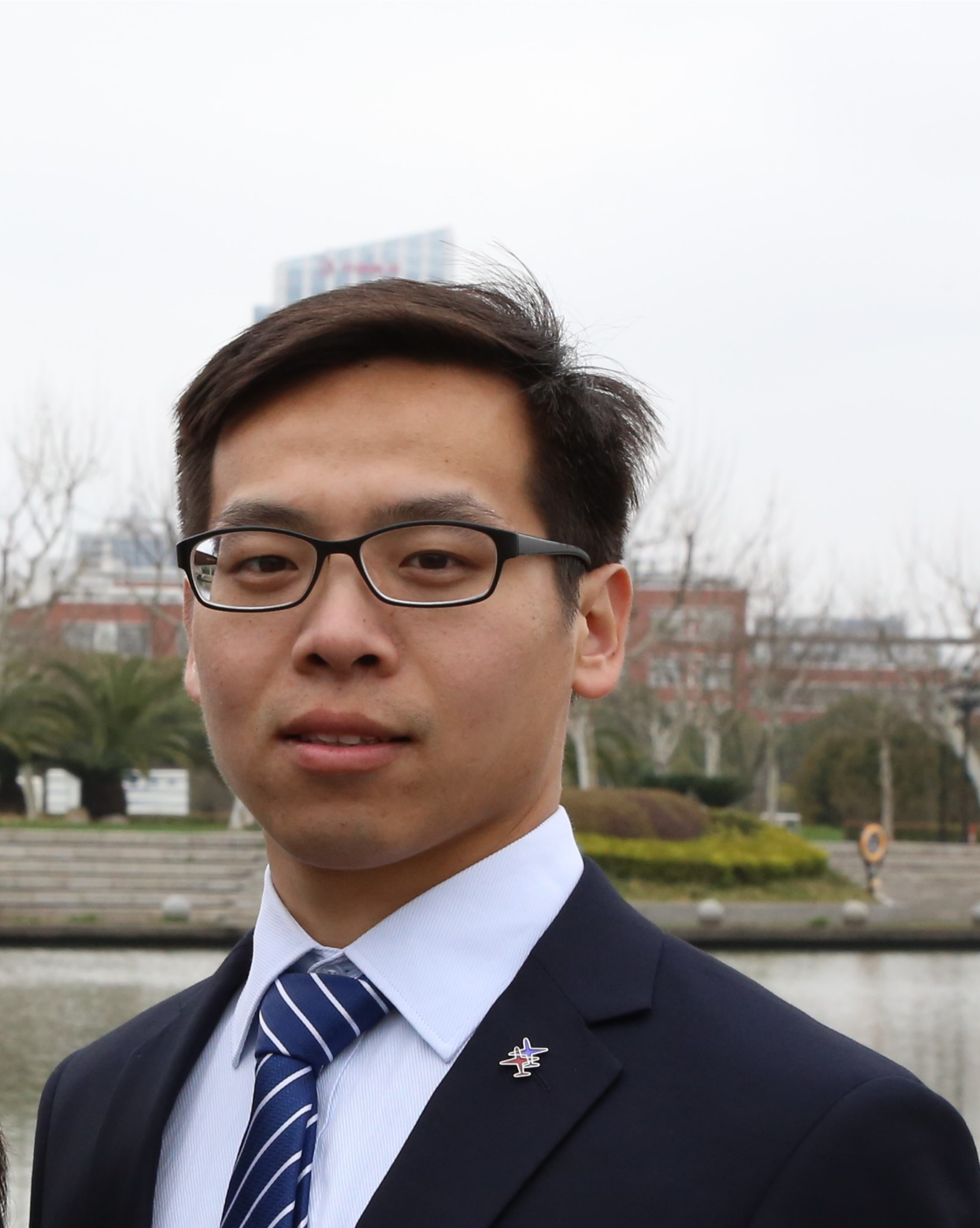}}]{Sensen Liu} received the B.S. degree in mechanical engineering from Tongji University, Shanghai, China, in 2016. He is currently a Ph.D. candidate in the School of mechanical and engineering, Shanghai Jiao Tong University, China. His research focuses on aerial manipulation, gripper design, planning and control of unmanned aerial vehicles manipulation system. 
\end{IEEEbiography}
\vspace{-50 pt} 
\begin{IEEEbiography}[{\includegraphics[width=1in,height=1.2in,clip]{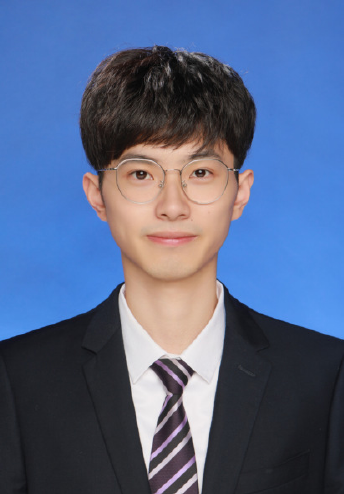}}]{Wenkang Hu} received a bachelor's degree in mechanical engineering from Shanghai Jiaotong University, Shanghai, China, 2021. Now he is a master candidate in the State Key Laboratory of mechanical system and vibration, Shanghai Jiaotong University.His research interests are multi-agent cluster cooperation and rapid flight decision-making of aerial robots. 
\end{IEEEbiography}
\vspace{-50pt} 
\begin{IEEEbiography}[{\includegraphics[width=1in,height=1.2in,clip]{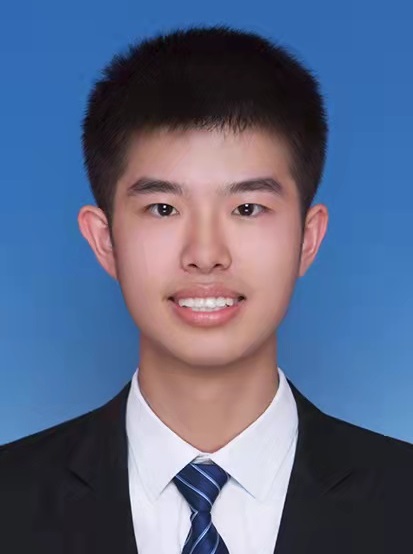}}]{Zhaoying Wang} received the B.E. degree in mechanical engineering from Wuhan University, Wuhan, China, in 2019. He is now a Ph.D. candidate at the State Key Laboratory of Mechanical System and Vibration, Shanghai Jiao Tong University. His research interest is multi-agent cooperative system and visual positioning of micro aerial vehicles.
\end{IEEEbiography}
\vspace{-50 pt} 
\begin{IEEEbiography}[{\includegraphics[width=1in,height=1.2in,clip]{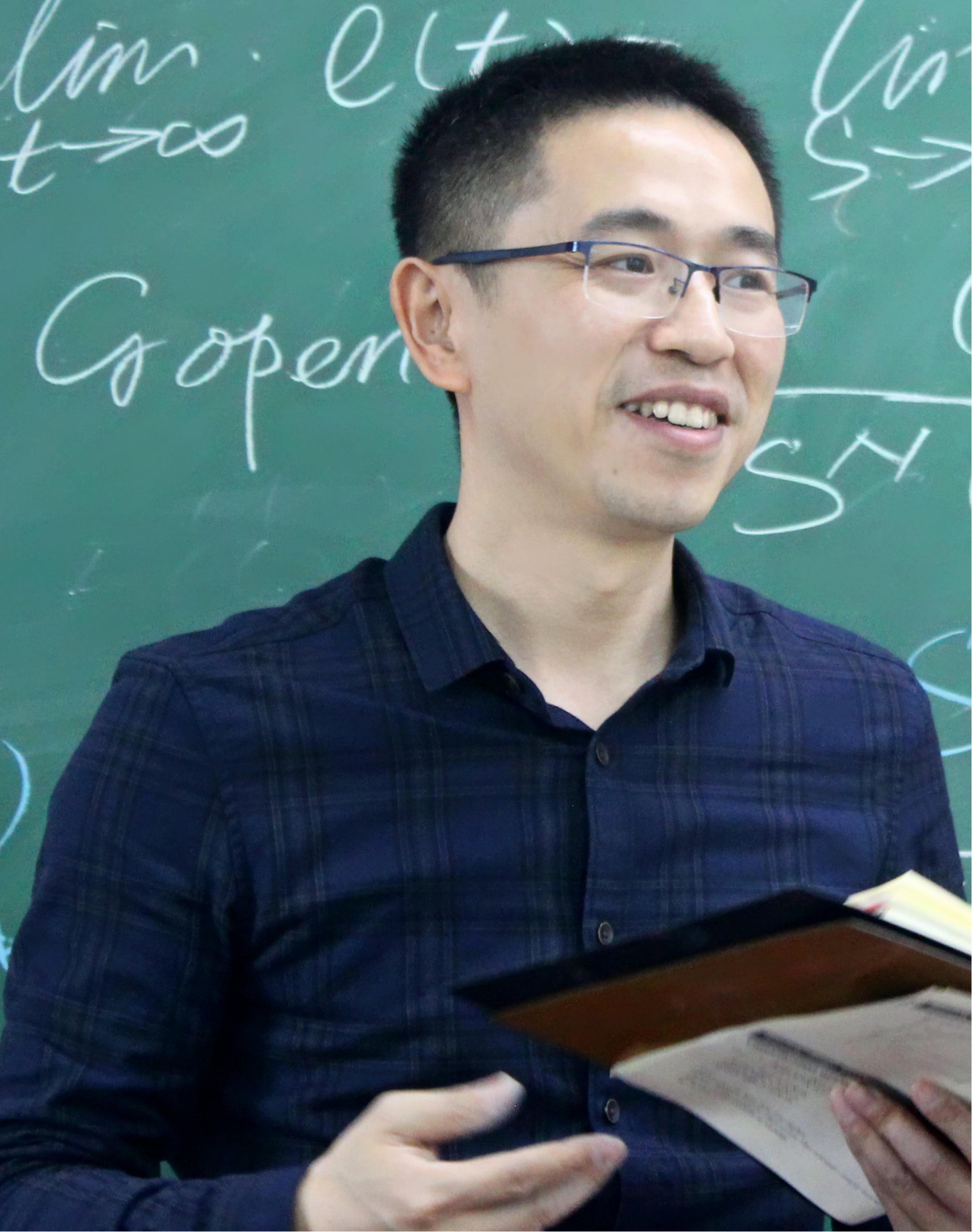}}]{Wei Dong} received the B.S. degree and Ph.D. degree in mechanical engineering from Shanghai Jiao Tong University,
Shanghai, China, in 2009 and 2015, respectively. He is currently an associate professor in the Robotic Institute, School of Mechanical Engineering, Shanghai Jiao Tong University. For years, his research group was champions in several national-wide autonomous navigation
competitions of unmanned aerial vehicles in China. In 2022, he was selected into the Shanghai Rising-Star Program for distinguished young scientists. His research interests include cooperation perception and agile control of unmanned systems.
\end{IEEEbiography}
\vspace{-40 pt} 
\begin{IEEEbiography}[{\includegraphics[width=1in,height=1.2in,clip]{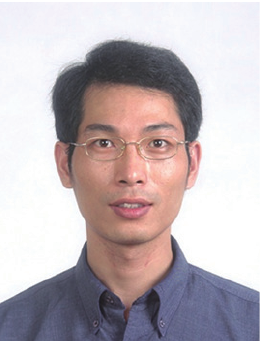}}]{Xinjun Sheng} received the B.Sc., M.Sc.,and Ph.D. degrees in mechanical engineering from Shanghai Jiao Tong University, Shanghai, China, in 2000, 2003, and 2014, respectively. He is currently a Professor in the School of Mechanical Engineering, Shanghai Jiao Tong University. His current research interests include robotics, and bio-mechatronics. Dr. Sheng is a Member of the IEEERAS, the IEEEEMBS,and the IEEEIES.
  \end{IEEEbiography}




\end{document}